\documentclass{article}

\usepackage[preprint]{neurips_2025}

\usepackage[utf8]{inputenc} % allow utf-8 input
\usepackage[T1]{fontenc}    % use 8-bit T1 fonts
\usepackage[hidelinks]{hyperref}
\usepackage{url}            % simple URL typesetting
\usepackage{booktabs}       % professional-quality tables
\usepackage{amsfonts}       % blackboard math symbols
\usepackage{nicefrac}       % compact symbols for 1/2, etc.
\usepackage{microtype}      % microtypography
\usepackage{xcolor}         % colors
\usepackage{graphicx}
\usepackage{comment}
\usepackage{multicol}
\usepackage{multirow}
\title{JEPA-DNA: Grounding Genomic Foundation Models through Joint-Embedding Predictive Architectures}

\author{%
\href{https://orcid.org/0000-0002-5006-9300}{\includegraphics[scale=0.06]{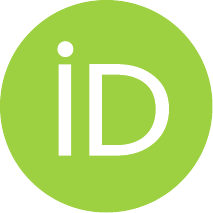}\hspace{1mm}Ariel Larey$^{1}$}
\And
\href{https://orcid.org/0009-0009-5635-6763}{\includegraphics[scale=0.06]{orcid.pdf}\hspace{1mm}Elay Dahan$^{2}$}
\And
\href{https://orcid.org/0009-0004-8952-3672}{\includegraphics[scale=0.06]{orcid.pdf}\hspace{1mm}Amit Bleiweiss$^{3}$}
\And
\href{https://orcid.org/0009-0000-0029-342X}{\includegraphics[scale=0.06]{orcid.pdf}\hspace{1mm}Raizy Kellerman$^{4}$}
\And
\href{https://orcid.org/0009-0007-7314-7934}{\includegraphics[scale=0.06]{orcid.pdf}\hspace{1mm}Guy Leib$^{4}$}
\And
\href{https://orcid.org/0000-0001-5060-891X}{\includegraphics[scale=0.06]{orcid.pdf}\hspace{1mm}Omri Nayshool$^{4}$}
\And
\href{https://orcid.org/0000-0001-5136-8014}{\includegraphics[scale=0.06]{orcid.pdf}\hspace{1mm}Dan Ofer$^{4}$}
\And
\href{https://orcid.org/0000-0002-5580-5282}{\includegraphics[scale=0.06]{orcid.pdf}\hspace{1mm}Tal Zinger$^{4}$}
\And
\href{https://orcid.org/0000-0003-1121-5130}{\includegraphics[scale=0.06]{orcid.pdf}\hspace{1mm}Dan Dominissini$^{4}$}\\
\And
\href{https://orcid.org/0000-0002-4594-1811}{\includegraphics[scale=0.06]{orcid.pdf}\hspace{1mm}Gideon Rechavi$^{4}$}\\
\And
\href{https://orcid.org/0000-0002-2609-4354}{\includegraphics[scale=0.06]{orcid.pdf}\hspace{1mm}Nicole Bussola$^{5}$}\\
\And
\href{https://orcid.org/0000-0003-1874-2515}{\includegraphics[scale=0.06]{orcid.pdf}\hspace{1mm}Simon Lee$^{5}$}\\
\And
\href{https://orcid.org/0000-0002-1693-9157}{\includegraphics[scale=0.06]
{orcid.pdf}\hspace{1mm}Shane O’Connell$^{5}$}\\
\And
\href{https://orcid.org/0000-0001-5951-1866}{\includegraphics[scale=0.06]
{orcid.pdf}\hspace{1mm}Dung Hoang$^{5}$}\\
\And
\href{https://orcid.org/0000-0001-6488-6240}{\includegraphics[scale=0.06]{orcid.pdf}\hspace{1mm}Marissa Wirth$^{5}$}\\
\And
\href{https://orcid.org/0000-0001-8135-6858}{\includegraphics[scale=0.06]{orcid.pdf}\hspace{1mm}Alexander W. Charney$^{5}$}\\
\And
\href{https://orcid.org/0000-0002-0939-3379}{\includegraphics[scale=0.06]{orcid.pdf}\hspace{1mm}Nati Daniel$^{1,\ast}$}
\href{https://orcid.org/0000-0002-3393-6070}{\includegraphics[scale=0.06]{orcid.pdf}\hspace{1mm}Yoli Shavit$^{1,}$}\thanks{%
Co-corresponding authors: \texttt{ndaniel@nvidia.com}, \texttt{yolis@nvidia.com}.
$^{1}$Applied AI Architecture, NVIDIA, Israel. 
$^{2}$Worldwide Field Ops, NVIDIA, Israel. 
$^{3}$Developer Programs, NVIDIA, Israel. 
$^{4}$Cancer Research Center and Wohl Institute of Translational Medicine, Sheba Medical Center, Tel Hashomer, Israel. 
$^{5}$Windreich Department of AI and Human Health, Icahn School of Medicine at Mount Sinai, New York, USA.
}\\
}

\begin{document}
\maketitle
\begin{abstract}
Genomic Foundation Models (GFMs) typically rely on Masked Language Modeling (MLM) or Next-Token Prediction (NTP) to learn the "Laws of Nature". While effective at capturing local syntax, these generative paradigms prioritize token-level reconstruction over high-level functional context. We introduce JEPA-DNA, a model-agnostic continual training framework that integrates a Joint-Embedding Predictive Architecture (JEPA) with traditional generative objectives. By supervising global sequence embeddings in a latent space, JEPA-DNA forces models to predict the functional representations of masked genomic segments, shifting the learning signal from token recovery to semantic alignment. We evaluate JEPA-DNA on 17 diverse genomic benchmark tasks, demonstrating consistent gains in linear probing and zero-shot performance regardless of the underlying GFM architecture or generative objective. Our framework establishes a new state-of-the-art for GFMs, surpassing the best existing models by bridging generative precision with latent semantic grounding. Through extensive ablation studies, we further characterize the synergistic interplay between generative and latent objectives. Our code is publicly available at https://github.com/NVIDIA-Digital-Bio/JEPA-DNA.
\end{abstract}

\section{Introduction}\label{sec:introduction}
Genomic Foundation Models (GFMs) have significantly advanced our ability to model the complex interactions within DNA sequences \citep{benegas2025genomic}. Architectures such as DNABERT-2~\citep{zhou2024dnabert}, Nucleotide Transformer (NT)~\citep{dalla2025nucleotide}, HyenaDNA~\citep{nguyen2023hyenadna}, Evo~\citep{nguyen2024sequence} and Caduceus~\citep{schiff2024caduceus} adapt Large Language Models (LLMs) to genomic data, utilizing context windows from several kilobases to megabase scales over backbones that range from Transformers to long convolutions and bi-directional Mamba layers. Typically, these models rely on generative self-supervised objectives, such as Masked Language Modeling (MLM) or Next Token Prediction (NTP), to learn representational spaces.

% While effective at identifying local motifs and sequence patterns \citep{zhou2024dnabert, dalla2025nucleotide}, generative objectives focus heavily on token-level reconstruction. This localized supervision can lead models to over-allocate capacity to high-frequency sequence "noise" (e.g. repetitive elements or neutral polymorphisms) at the expense of global functional context. Consequently, GFMs may achieve high syntactic fidelity without internalizing the high-level regulatory logic or long-range interactions that govern genomic function. 

While effective at identifying local motifs and sequence patterns \citep{zhou2024dnabert, dalla2025nucleotide}, generative objectives focus heavily on token-level reconstruction. Motivated by joint-embedding predictive architectures, which suggest that reconstruction in input space can prioritize low-level details over more abstract semantic structure \citep{assran2023self}, we propose a framework designed to capture the high-level regulatory logic and long-range interactions that govern genomic function.

To bridge the gap between genomic syntax and biological semantics, we introduce JEPA-DNA, a novel framework that integrates the Joint-Embedding Predictive Architecture (JEPA)~\citep{lecun2022path} into genomic training. In JEPA, the predictor must recover the latent representation of a masked span from the surrounding context, which can only be satisfied by encoding sequence-level regularities rather than local token statistics. By coupling token-level recovery with latent-space prediction, JEPA-DNA encourages the learning of abstract, functional features invariant to low-level sequence noise. 

 We evaluate JEPA-DNA across 17 genomic benchmark tasks, utilizing linear probing and zero-shot protocols to isolate representational quality~\citep{larey2026gfmbench}. Our approach is uniquely modular, serving as a continual pre-training phase to refine existing GFMs across disparate architectures and objectives. Empirical results demonstrate that latent prediction consistently improves performance, establishing a new state-of-the-art for GFMs across supervised linear probing tasks. Extensive ablation studies confirm that performance gains are driven by JEPA-DNA's core components rather than simply additional training.

In summary, our contributions are as follows:
\begin{itemize}
    \item We introduce a novel continual training framework for GFMs that shifts the learning objective from literal token reconstruction to latent feature alignment via a Joint-Embedding Predictive Architecture. To the best of our knowledge, JEPA-DNA is the first to adapt Joint-Embedding Predictive Architectures to high-resolution, multiscale genomic sequences.
    \item We evaluate JEPA-DNA on 17 diverse genomic benchmark tasks, demonstrating consistent gains in linear probing and zero-shot performance across leading GFM architectures (DNABERT-2~\citep{zhou2024dnabert}, NTv3~\citep{boshar2025foundational}, and HyenaDNA~\citep{nguyen2023hyenadna}), tokenization methods, and generative objectives (MLM and NTP).
    \item We establish a new state-of-the-art under our frozen linear probing settings, across supervised tasks, empirically showing that JEPA-DNA learns more linearly-separable and biologically relevant features than standard generative baselines.
\end{itemize}
\section{Related Work}\label{sec:related_Work}
\subsection{Pre-training Paradigms of  Genomic Foundation Models}\label{subsec:gfm_related} 
The development of GFMs has been largely inspired by the success of LLMs in Natural Language Processing. Early iterations, such as DNABERT~\citep{ji2021dnabert}, adapted the BERT architecture~\citep{devlin2019bert} replacing its original WordPiece subword tokenizer with k-mer tokenization to capture bidirectional context within the genome. More recently, DNABERT-2~\citep{zhou2024dnabert} and the Nucleotide Transformer~\citep{dalla2025nucleotide} expanded this scale by training on diverse multi-species datasets, demonstrating that larger context windows and higher parameter counts can improve "out-of-the-box" performance on downstream tasks like promoter prediction and variant effect prediction.

Beyond Transformer-based architectures, recent advancements have focused on overcoming the quadratic scaling of self-attention to model longer genomic dependencies. HyenaDNA~\citep{nguyen2023hyenadna} utilizes the Hyena operator to process sequences at single-nucleotide resolution across long contexts. Caduceus~\citep{schiff2024caduceus} replaces attention with bi-directional Mamba (state-space) layers. Similarly, Evo2~\citep{brixi2026genome} leverages a StripedHyena 2 backbone to push the context size to megabase scales. In parallel, Nucleotide Transformer v3 (NTv3) ~\citep{boshar2025foundational} extends this line of work by wrapping a Transformer backbone within a U-Net–style architecture, enabling efficient downsampling and upsampling across megabase-scale genomic windows while maintaining single-nucleotide-resolution predictions. Despite these structural innovations, nearly all current GFMs rely on token-level reconstruction objectives, namely MLM and NTP. While these methods are effective for learning the "syntax" of DNA, they often fail to capture the structural "logic" or functional state of the genome~\citep{benegas2023dna}.

Alternative self-supervised strategies have sought to move beyond literal token reconstruction by incorporating evolutionary or contrastive constraints. Models such as GPN-MSA~\citep{benegas2025dna} leverage multi-sequence alignment (MSA) to learn from cross-species conservation patterns, identifying functional constraints through substitution probabilities. In parallel, contrastive learning frameworks like DNASimCLR~\citep{yang2024dnasimclr} employ data augmentations to learn representations by minimizing the distance between similar sequence "views" in latent space. However, these paradigms involve significant trade-offs: MSA-based methods require computationally expensive preprocessing and are limited by the availability of high-quality alignments, while contrastive methods often necessitate non-trivial augmentations or costly negative mining. In contrast, JEPA-DNA captures functional invariants directly from the raw sequence.

\subsection{Joint-Embedding Predictive Architecture (JEPA)}
\label{subsec:jepa_related} 
In computer vision, the Joint-Embedding Predictive Architecture (JEPA) has emerged as a powerful alternative to generative modeling. In this paradigm, a model is tasked with predicting the latent representation of a masked segment rather than its literal pixels~\citep{assran2023self}. This shift from signal reconstruction to representation prediction allows the model to ignore unpredictable, low-level details and focus on semantically rich features. This concept was recently extended to the language domain with LLM-JEPA~\citep{huang2026llm}, which proposes a first step towards coupling token-level recovery with sequence-level latent grounding for natural language tasks. Within the biological domain, the JEPA framework was recently introduced by GeneJEPA for single-cell transcriptomics~\citep{litman2025genejepa}. However, GeneJEPA operates on gene-expression vectors, designed to learn and reason about gene-gene relationships within a cell. Unlike transcriptomic models that process tabular gene sets, our framework must handle the multiscale nature of DNA, where functional meaning is encoded through sequential motifs and long-range dependencies.
\section{Method}\label{sec:method}
The JEPA-DNA framework treats the genome not merely as a sequence of tokens, but as a structured signal with both local syntax and global semantics. We achieve this by augmenting a standard GFM backbone $\mathcal{E}$ with a Joint-Embedding Predictive Architecture (JEPA) branch, enabling a multi-objective learning process.

\subsection{Architecture Components}
\begin{figure}
    \centering
    \includegraphics[scale=0.35]{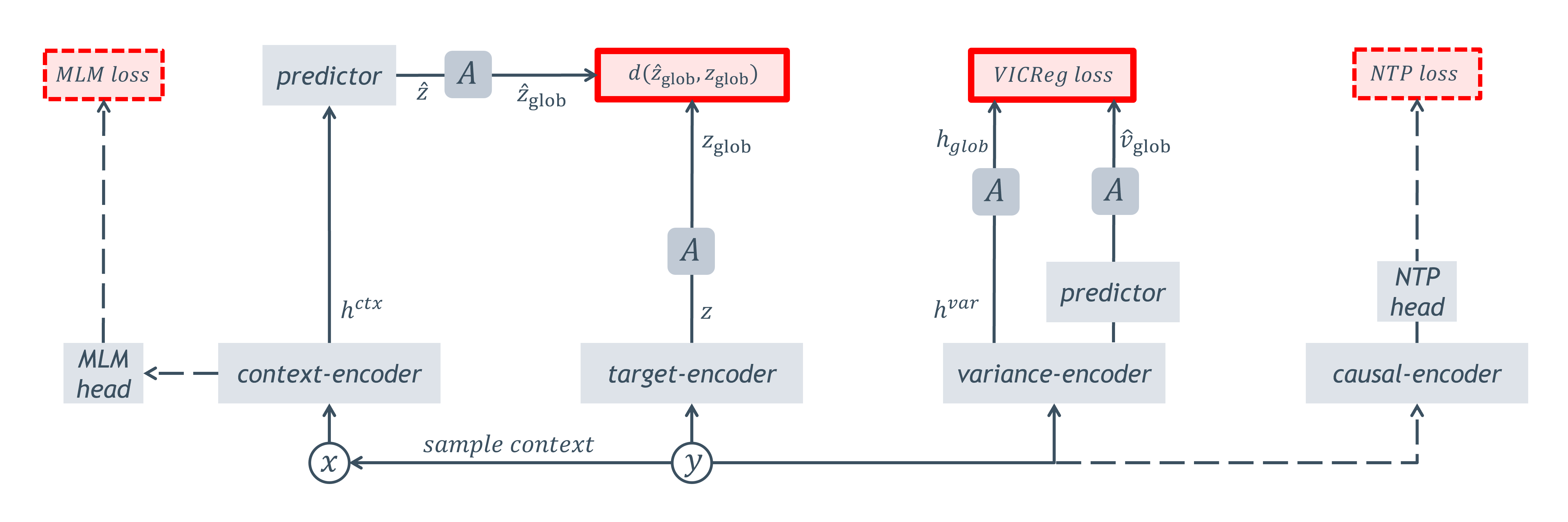}
    \caption{The JEPA-DNA Architecture.}
    \label{fig:arch}
\end{figure}

The JEPA-DNA architecture is illustrated in Fig.~\ref{fig:arch}. Its core elements are as follows:
\begin{itemize}
    \item \textbf{Context-encoder ($\mathcal{E}_{\theta}$):} A GFM backbone that processes the masked input sequence $\mathbf{x}$ and produces a contextual latent sequence $\mathbf{h}^{ctx} = \mathcal{E}_{\theta}(\mathbf{x})$.
    \item \textbf{Target-encoder ($\mathcal{E}_{\bar{\theta}}$):} A structural duplicate of the context encoder, whose weights $\bar{\theta}$ are updated via an exponential moving average (EMA) of $\theta$. This encoder processes the corresponding unmasked target sequence $\mathbf{y}$ to yield a latent target sequence $\mathbf{z} = \mathcal{E}_{\bar{\theta}}(\mathbf{y})$.
    \item \textbf{Predictor head ($\mathcal{P}_{\phi}$):} A network designed to map context representations into the target latent space. It takes the contextual sequence $\mathbf{h}^{ctx}$ as input and outputs predicted target latents $\hat{\mathbf{z}} = \mathcal{P}_{\phi}(\mathbf{h}^{ctx})$, thereby learning to align context-derived representations with the target encoder’s latent sequence.
    \item \textbf{Sequence aggregator ($\mathcal{A}$):} An aggregation operator that maps latent sequence embeddings to a single global sequence representation, instantiated according to the backbone’s native behavior (e.g., $[CLS]$ token embedding, last token embedding, or mean pooling). Given a target latent sequence $\mathbf{z}$, it produces a global target representation $\mathbf{z}_{\mathrm{glob}} = \mathcal{A}(\mathbf{z})$, and when applied to the predictor outputs $\hat{\mathbf{z}}$ it yields the corresponding predicted global latent representation $\hat{\mathbf{z}}_{\mathrm{glob}} = \mathcal{A}(\hat{\mathbf{z}})$.
\end{itemize}

In addition, we introduce a \textbf{causal-encoder} for auto-regressive backbones (as detailed in Section \ref{llm_loss}) and an additional \textbf{variance-encoder} to address mode collapse (as detailed in Section \ref{vicreg_loss}).

\subsection{Masking and Re-masking Strategy}\label{subsec:masking}
The JEPA-DNA framework relies on a dual-masking process, to ensure the predictor does not see the masked content.  
% \paragraph{Initial Masking.} Given an input sequence, a subset of tokens is replaced by a special $[MASK]$ token. This masked sequence is subsequently processed by the context encoder $\mathcal{E}_{\theta}$ to yield latent representations. In contrast to standard Masked Language Modeling (MLM) protocols that typically mask random independent tokens (approximately 15\%), our approach employs a span-based masking strategy. We sample multiple contiguous target regions rather than individual tokens, resulting in a higher aggregate masking ratio (typically exceeding 20\%), in alignment with the JEPA context encoder masking protocol in vision~\citep{assran2023ijepa}. In particular, this masking configuration is propagated to both the JEPA loss and the standard LLM loss, thereby enforcing a uniformly more challenging reconstruction objective. 

\textbf{Initial Masking.} Given an input sequence $\mathbf{y}$, a subset of tokens is replaced by special $[MASK]$ tokens, producing $\mathbf{x}$. This masked sequence is subsequently processed by the context encoder $\mathcal{E}_{\theta}$ to yield latent representations $\mathbf{h}^{ctx}$. In contrast to standard MLM protocols, which typically mask random, independent tokens (approximately 15\%), our approach adopts a span-based masking strategy. We sample multiple contiguous target regions rather than individual tokens, resulting in a higher aggregate masking ratio (typically exceeding 20\%), in alignment with the JEPA context encoder masking protocol in vision~\citep{assran2023self}. Furthermore, we introduce a masking scheduler that gradually increases both the masking ratio and the number of masked spans over the course of training. Early in training, the model is exposed to relatively mild corruption patterns, while later stages employ more aggressive masking configurations, progressively challenging the model to infer longer-range dependencies from increasingly limited context. In particular, this masking configuration is propagated to both the JEPA loss and the standard MLM loss, thereby enforcing a uniformly more challenging reconstruction objective.

\textbf{Re-masking Strategy.} To prevent the predictor from having a trivial mapping to the targets, we introduce a re-masking step, where $\mathbf{h}^{ctx}$, the outputs of $\mathcal{E}_{\theta}$, at the masked positions are replaced by a learnable latent $[MASK]$ embedding. Note that the predictor head receives the full sequence of context encodings. By conditioning the predictor on the entire encoded sequence rather than a single vector, the model can leverage both localized spatial information and the global sequence summary to accurately reconstruct the target latent $\hat{\mathbf{z}}_{\mathrm{glob}}$. 
%\subsection{Training Pipeline}
%JEPA-DNA training proceeds in three stages. In the first stage, each genomic backbone model undergoes pre-training with its native generative objective on large-scale genomic data, using masked language modeling for bidirectional encoders or next-token prediction for autoregressive models. In the second stage, JEPA-based continual training is performed on the same pre-training dataset while the backbone encoder parameters are held fixed and only an attached predictor branch is optimized. In the third stage, JEPA-based training is continued, but both the encoder and the predictor are updated jointly under a combined objective.
\vspace{-0.5em}
\subsection{Multi-Objective Continual training}
JEPA-DNA employs a two-stage continual training procedure starting from a GFM backbone previously pre-trained on large-scale genomic data using its native objective (MLM or NTP). During the first stage, the backbone encoder remains frozen while only the attached predictor branch is optimized. In the second stage, the encoder and predictor are updated jointly, refining the model under the combined JEPA-DNA objective.

The model is optimized by minimizing a composite loss function that balances language modeling, latent prediction, and embedding diversity.
% \subsubsection{LLM Loss ($\mathcal{L}_{llm}$) }
% To maintain nucleotide-level precision, we retain the standard LLM objective (MLM or NTP). For a masked sequence with indices $\mathcal{M}$, the loss is:
% \begin{equation}
%     \mathcal{L}_{llm} = -\sum_{i \in \mathcal{M}} \log P(x_i | \mathbf{h}_i)
% \end{equation}
% where $\mathbf{h}_i$ denotes the hidden state of the genomic token, and the set of masked indices $\mathcal{M}$ is derived from the masking strategy.

\subsubsection{LLM Loss ($\mathcal{L}_{llm}$) }
\label{llm_loss}
To maintain token-level precision, we retain the standard LLM objective (MLM or NTP). For a masked sequence with indices $\mathcal{M}$, the loss is:

\begin{equation}
\mathcal{L}_{llm} = -\sum_{i \in \mathcal{M}} \log P(t_i \mid h_i)
\end{equation}

where $P(t_i)$ denotes the predicted probability of the genomic token $t_i$ at position $i$ given the contextual hidden state $h_i$. The set of indices $\mathcal{M}$ is derived from the masking strategy in the MLM setting or from the causal autoregressive formulation in the NTP setting. In the latter case, the objective is computed using an additional forward pass over the full sequence through a causal-encoder that shares its weights with the context encoder.

\subsubsection{Latent Predictive Loss ($\mathcal{L}_{jepa}$)}
The JEPA objective grounds the model by forcing the global representation $\mathbf{h}_{\mathrm{glob}}$ to capture functional semantics. The predictor's global head $\hat{\mathbf{z}}_{\mathrm{glob}}$ attempts to match the embedding produced by the target encoder $\mathbf{z}_{\mathrm{glob}}$ in the latent space. 

We define this loss using cosine similarity:
\begin{equation}
\mathcal{L}{jepa} = 1 - \frac{\hat{\mathbf{z}}_{\mathrm{glob}} \cdot \mathbf{z}_{glob}}{||\hat{\mathbf{z}}_{\mathrm{glob}}||_2 \cdot ||\mathbf{z}_{glob}||_2}
\end{equation}
 Minimizing this objective encourages the predicted representation of the context encoder to align with the functional embedding of the target encoder. %Unlike $L_2$ minimization, which can be sensitive to absolute feature scales, cosine similarity ensures that the model learns the relative semantic relationships and functional states inherent in the genomic sequence.

% \subsubsection{Variance Regularization ($\mathcal{L}_{var}$)}
% To prevent the "collapse" problem common in non-contrastive methods (where the model outputs a constant vector), we utilize a variance constraint on the latent vectors $\mathbf{Z}$ across a batch:
% \begin{equation}
%     \mathcal{L}_{var} = \frac{1}{d} \sum_{j=1}^{d} \max(0, \gamma - \sigma(\mathbf{z}_{:,j}))
% \end{equation}
% where $\sigma(\mathbf{z}_{:,j})$ is the standard deviation of the $j$-th dimension across the batch, and $\gamma$ is a constant threshold (typically $\gamma = 1$).

\subsubsection{Variance and Covariance Regularization}
\label{vicreg_loss}
To prevent the ``collapse'' problem common in non-contrastive methods (where the 
model outputs a constant vector), we utilize variance and covariance constraints 
on the latent vectors $\mathbf{h}_{\mathrm{glob}}, \hat{\mathbf{v}}_{\mathrm{glob}} \in \mathbb{R}^{B \times d}$ across a batch of 
size $B$ and hidden dimension $d$, following the VICReg framework \citep{bardes2021vicreg}. 
These vectors are obtained via a separate forward pass through a deterministic variance encoder 
(without dropouts or masking) that produces $\mathbf{h}^{\mathrm{var}}$ and shares its weights with the context encoder. 
To keep the setup consistent with the JEPA objective, the VICReg branch also employs a deterministic predictor head (no dropouts) whose architecture and weights are tied to those of the main JEPA predictor.

\paragraph{Variance Loss ($\mathcal{L}_{var}$).}
The variance loss ensures that each embedding dimension maintains sufficient 
variance across the batch, preventing informational collapse:
\begin{equation}
    \mathcal{L}_{var} = \frac{1}{d} \sum_{j=1}^{d} \max(0, \gamma - \sigma(\mathbf{u}_{:,j}))
\end{equation}
where $\sigma(\mathbf{u}_{:,j})$ is the standard deviation of the $j$-th 
dimension across the batch, and $\gamma$ is a constant threshold (typically 
$\gamma = 1$).

\paragraph{Covariance Loss ($\mathcal{L}_{cov}$).}
The covariance loss encourages decorrelation between different embedding 
dimensions, preventing redundant representations:
\begin{equation}
    \mathcal{L}_{cov} = \frac{1}{d} \sum_{i \neq j} C_{ij}^2
\end{equation}
where $C \in \mathbb{R}^{d \times d}$ is the covariance matrix of the 
centered embeddings:
\begin{equation}
    C = \frac{1}{B-1} (\mathbf{U} - \bar{\mathbf{U}})^\top (\mathbf{U} - \bar{\mathbf{U}})
\end{equation}
and $\bar{\mathbf{U}}$ denotes the batch mean. By penalizing the squared 
off-diagonal elements of the covariance matrix, this loss encourages each 
dimension to encode independent information.

\subsubsection{Total Objective}
The final optimization problem is defined as:
\begin{equation}
    \min_{\theta, \phi} \mathcal{L}_{total} = \lambda_1 \mathcal{L}_{llm} + \lambda_2 \mathcal{L}_{jepa} + \lambda_3 \mathcal{L}_{var} + \lambda_4 \mathcal{L}_{cov}
\end{equation}
where $\lambda_{1,2,3,4}$ are hyper-parameters that weigh the contribution of each objective. 
\subsection{Compatibility across Model Architectures and Objectives}
The JEPA-DNA framework is model-agnostic and compatible with MLM and NTP objectives. Its primary requirement is an aggregation operator that compresses a sequence into a latent representation. While the $[CLS]$ token is “native” to the Transformer encoder architecture trained with MLM objectives~\citep{devlin2019bert}, JEPA-DNA also extends naturally to Transformer decoders, State Space Models (SSMs), and long-convolution backbones (e.g., HyenaDNA~\citep{nguyen2023hyenadna}) that are trained with NTP objectives. In these architectures, we use the embeddings of the last $[SEP]$ token of the sequence, and the same JEPA formulation is equally applicable when the aggregation operator is implemented as mean pooling over token embeddings, as in Nucleotide Transformer–style models~\citep{boshar2025foundational}. By supervising the DNA sequence's global representation within the JEPA loss, regardless of the generative objective or specific encoder architecture, we enforce a global pooling mechanism, effectively compensating for the otherwise local or recurrent nature of these operators.
\section{Experimental Results}\label{sec:results}
We evaluate the JEPA-DNA framework across diverse architectures and benchmarks to quantify its impact on representational quality. 
\subsection{Experimental Setup}\label{subsec:exp_setup}
We provide the core methodological and implementation details below, while deferring extended specifications, training and implementation details to the Appendix (Section~\ref{subsec:appendix_impl_details}).

\paragraph{Backbones \& Predictor Architecture.} To evaluate the compatibility of JEPA-DNA, we select three representative models to serve at its encoder backbone, spanning the dominant architectural paradigms in GFM research: (1) DNABERT-2~\citep{zhou2024dnabert}, a 117M-parameter Transformer Encoder utilizing MLM and BPE tokenization; (2) NTv3~~\citep{boshar2025foundational}, a 100M-parameter hybrid U-Net/Transformer model utilizing MLM over single nucleotides; and (3) HyenaDNA~\citep{nguyen2023hyenadna}, a sub-quadratic operator model optimized via an NTP objective. These models vary in their sequence aggregation strategies, utilizing $[CLS]$ embeddings, mean pooling, and last tokens, respectively. The JEPA-DNA framework introduces a target encoder (EMA-updated version of the GFM backbone) and a lightweight predictor. For token-aggregated backbones (DNABERT-2, HyenaDNA), the predictor is a 4-layer Transformer that reconstructs the target’s latent representations in a reduced embedding space. For mean-pooled backbones (NTv3), we employ a 3-layer MLP predictor that operates directly on the pooled sequence embedding. This dual-design ensures that the latent prediction objective is compatible with both local attention-based and global pooling-based architectures.

\paragraph{Baselines.} {In addition to the original pre-trained model (\emph{original checkpoint}), we report two continually-trained baselines to isolate the original checkpoint selection, and separate the effect of continued optimization from that of the latent objective. Both continue training each backbone with the same native objective and dataset used for its own pre-training, and differ only in budget: the \emph{step-matched} baseline is trained for the same number of iterations as JEPA-DNA, whereas the \emph{compute-matched} baseline is trained for the same amount of compute time (see Appendix Table~\ref{tab:cost} for compute details).

\paragraph{Continual Pre-training Datasets and Evaluation Benchmarks.} To ensure that performance improvements are attributable solely to the JEPA-DNA objective rather than exposure to novel data, we perform continual training using subsets of the original pre-training corpora for each backbone. For DNABERT-2, we utilize the multi-species corpus (human and five model organisms, 4.76M sequences); for NTv3, we use the Open Genome 2 corpus~\citep{brixi2026genome} (random crops of 8,192 bp); and for HyenaDNA, we utilize the hg38 human reference (600k windows). By maintaining data parity with the original baselines, we isolate the impact of our latent-prediction framework as a general "representational uplift" mechanism. We utilize GFMBench-API~\citep{larey2026gfmbench} to evaluate representational quality across a standardized suite of genomic tasks. Our supervised evaluation includes nine tasks spanning regulatory syntax (GUE~\citep{zhou2024dnabert}) and functional variants (VariantBenchmarks~\citep{medvedev2025biotoken}, Long Range Benchmark (LRB)~\citep{kao2024advancing}). Our zero-shot evaluation covers eight clinical and fitness-scoring benchmarks tasks from BEND~\citep{marin2024bend}, TraitGym~\citep{benegas2025benchmarking}, BRCA1~\citep{findlay2018accurate}, and ClinVar-based suites to measure emergent biological priors ~\citep{benegas2025dna}.

%\paragraph{Evaluation Protocol.} We assess representation quality through two paradigms: linear probing and zero-shot fitness scoring. For linear probing, we evaluate the latent space's linear separability by training a single-layer classifier on a frozen backbone, measuring performance via Area Under the Receiver Operating haracteristic curve (AUROC), Area Under the Precision-Recall Curve (AUPRC) and (3) Matthews Correlation Coefficient (MCC). In zero-shot evaluation, we measure emergent biological priors and semantic understanding by computing the cosine similarity between reference and mutant latent representations without any task-specific updates.
%\paragraph{Implementation and Training Schedule.} JEPA-DNA is implemented in PyTorch and optimized via SGDM with a momentum of 0.9. We introduce a specialized two-phase training schedule to stabilize the alignment of pre-trained embeddings: (i) a Predictor Warmup (1,000 steps) where the backbone is frozen to allow the predictor to ground its initial mappings, followed by (ii) Joint Training, where the backbone and predictor are optimized together using a low-learning-rate cosine decay. To prevent representation collapse, a common failure mode in joint-embedding architectures, we utilize an EMA target encoder with momentum ramping from 0.996 to 1.0. For the VicReg, we compute variance in evaluation mode using fixed-length sequences to ensure that the regularization reflects true semantic diversity rather than stochastic dropout or padding artifacts.

\subsection{Linear Evaluation}
Since JEPA-DNA acts during continual pre-training, the quantity of interest is the representation it produces. To evaluate the linear separability of functional features, we perform linear probing by training a single-layer classifier on top of a frozen backbone.

%Our evaluation spans disparate backbone architectures, training objectives, and aggregation strategies, using three complementary metrics: Area Under the Receiver Operating Characteristic curve (AUROC), Area Under the Precision-Recall Curve (AUPRC), and the Matthews Correlation Coefficient (MCC).

\begin{table}[tbh!]
\centering
\scriptsize
\renewcommand{\arraystretch}{1.1}
\caption{Linear probing performance (AUROC) across diverse GFM backbones, objectives, aggregation strategies, and self-supervised datasets (DS). Gain denotes the \emph{relative} AUROC gain of JEPA-DNA over the original checkpoint / step-matched / compute-matched models. Bold values indicate the top performance within each backbone group. '$\dagger$' symbol indicates overall state-of-the-art under linear probing.}
\label{tab:supervised_main}
{\color{black}
%\resizebox{\textwidth}{!}{
\begin{tabular}{@{}l cccc r@{}}
\toprule
\textbf{Task} & \textbf{Original} & \textbf{Step-} & \textbf{Compute-} & \textbf{JEPA-DNA} & \textbf{Gain} \\
              & \textbf{checkpoint} & \textbf{matched} & \textbf{matched} & \textbf{(Ours)} & \scriptsize \% AUROC \\
\midrule
\multicolumn{6}{l}{\textit{DNABERT-2 (Transformer + MLM + CLS; Multi-Species DS)}} \\
\midrule
GUE TF Binding    & 0.788 & 0.793 & 0.796 & \textbf{0.839} & +6.47 / +5.81 / +5.42 \\
GUE Promoter      & 0.884 & 0.927 & 0.927 & \textbf{0.928} & +4.98 / +0.06 / +0.11 \\
GUE Splice Site   & 0.653 & 0.670 & 0.672 & \textbf{0.705} & +7.96 / +5.28 / +4.87 \\
VB Coding Path.   & \textbf{0.624} & 0.622 & 0.605 & 0.610 & -2.24 / -1.88 / +0.77 \\
VB Exp. Effect    & 0.674 & 0.667 & 0.671 & \textbf{0.689}$^\dagger$ & +2.23 / +3.34 / +2.73 \\
VB Common/Rare    & \textbf{0.505} & 0.500 & 0.502 & 0.504 & -0.20 / +0.88 / +0.31 \\
VB meQTL          & 0.623 & 0.607 & 0.606 & \textbf{0.645}$^\dagger$ & +3.53 / +6.25 / +6.45 \\
VB sQTL           & \textbf{0.585} & 0.583 & 0.582 & 0.577 & -1.37 / -1.01 / -0.93 \\
LRB Causal eQTL   & 0.697 & 0.702 & 0.705 & \textbf{0.711}$^\dagger$ & +2.01 / +1.35 / +0.81 \\
\midrule
\multicolumn{6}{l}{\textit{Nucleotide Transformer v3 (Hybrid + MLM + Mean Pooling; Open-Genome 2 DS)}} \\
\midrule
GUE TF Binding    & 0.837 & 0.838 & 0.838 & \textbf{0.849}$^\dagger$ & +1.43 / +1.33 / +1.33 \\
GUE Promoter      & 0.907 & 0.907 & 0.907 & \textbf{0.937}$^\dagger$ & +3.31 / +3.30 / +3.30 \\
GUE Splice Site   & 0.713 & 0.713 & 0.712 & \textbf{0.733}$^\dagger$ & +2.81 / +2.87 / +2.89 \\
VB Coding Path.   & 0.668 & 0.667 & 0.667 & \textbf{0.675}$^\dagger$ & +1.05 / +1.17 / +1.16 \\
VB Exp. Effect    & 0.642 & 0.642 & 0.642 & \textbf{0.654} & +1.87 / +1.86 / +1.86 \\
VB Common/Rare    & 0.502 & 0.502 & 0.502 & \textbf{0.510} & +1.59 / +1.51 / +1.50 \\
VB meQTL          & 0.576 & 0.576 & 0.576 & \textbf{0.615} & +6.77 / +6.83 / +6.83 \\
VB sQTL           & 0.591 & \textbf{0.591} & 0.591 & 0.590 & -0.17 / -0.20 / -0.20 \\
LRB Causal eQTL   & 0.690 & 0.690 & 0.690 & \textbf{0.697} & +1.01 / +1.04 / +1.05 \\
\midrule
\multicolumn{6}{l}{\textit{HyenaDNA (Long Convolutions + NTP + last-token; Reference-Genome DS)}} \\
\midrule
GUE TF Binding    & 0.698 & \textbf{0.699} & 0.696 & 0.638 & -8.60 / -8.78 / -8.31 \\
GUE Promoter      & 0.686 & 0.689 & 0.699 & \textbf{0.763} & +11.22 / +10.77 / +9.18 \\
GUE Splice Site   & 0.506 & 0.509 & 0.515 & \textbf{0.558} & +10.28 / +9.53 / +8.37 \\
VB Coding Path.   & 0.515 & 0.514 & 0.510 & \textbf{0.559} & +8.54 / +8.82 / +9.54 \\
VB Exp. Effect    & 0.580 & \textbf{0.582} & 0.581 & 0.547 & -5.69 / -6.02 / -5.85 \\
VB Common/Rare    & 0.503 & 0.502 & 0.503 & \textbf{0.504} & +0.20 / +0.37 / +0.18 \\
VB meQTL          & 0.503 & 0.505 & 0.505 & \textbf{0.518} & +2.98 / +2.63 / +2.64 \\
VB sQTL           & 0.550 & 0.552 & \textbf{0.553} & 0.552 & +0.36 / +0.05 / -0.21 \\
LRB Causal eQTL   & 0.565 & 0.562 & 0.563 & \textbf{0.623} & +10.27 / +10.95 / +10.68 \\
\bottomrule
\end{tabular}
}%}
\end{table}

The results in Table~\ref{tab:supervised_main} demonstrate that JEPA-DNA is compatible with each of the evaluated backbone families and improves most tasks in all three (additional metrics are reported in Appendix Table~\ref{tab:supervised_detailed}). We observe consistent performance gains across all models, with JEPA-DNA improving performance in 6-8 out of 9 tasks for every backbone evaluated, regardless of the original pre-training paradigm (MLM or NTP) or sequence aggregation strategy (e.g., $[CLS]$, last token, or mean pooling). Notably, DNABERT-2, NTv3, and HyenaDNA exhibit AUROC gains of up to +7.96\%, +6.77\%, and +11.22\%, respectively, alongside mean improvements across all tasks of +2.60\%, +2.19\%, and +3.29\% compared to the original checkpoint, indicating that the latent-prediction objective reorganizes the embedding space to prioritize functional sequence properties across diverse genomic regimes.

The two continually-trained baselines further indicate that these gains are not a by-product of additional optimization. Continuing to train each backbone with its own native objective is not consistently beneficial: the step-matched and compute-matched baselines each improve only 15 of 27 task--backbone pairs, averaging $+0.17\%$ and $+0.22\%$ respectively, helping some tasks while degrading others. JEPA-DNA, in contrast, surpasses each of them on 22 of 27 comparisons, with mean relative gains of $+2.52\%$ and $+2.46\%$.

Across all evaluated backbones, JEPA-DNA achieves consistent performance gains in the majority of tasks {(6-8 out of 9, depending on the backbone)}, with substantial improvements in regulatory and structural identification tasks, including {promoter classification, splice site identification, meQTL and causal eQTL prediction}. We hypothesize that these gains arise because such tasks depend primarily on a broader regulatory context, making them well-suited to the global, context-focused objective of JEPA-DNA rather than to the reliance on local sequence variation. In particular, this interpretation is consistent with prior work showing that meQTL prediction and causal eQTL tasks capture how genetic variants interact with global attributes such as transcription factor binding, histone modifications, and chromatin accessibility \citep{ banovich2014methylation, wang2013transcription}
whereas tasks such as coding pathogenicity and sQTL prediction are often driven by more fine-grained, local sequence changes \citep{ zeng2022predicting}. The stability of these improvements is further validated through a multi-seed analysis (see Table~\ref{tab:supervised_full_seeds} in Appendix~\ref{subsec:appendix_robustness}).

We further observe that larger, higher-performing models tend to benefit more consistently from JEPA-DNA. For example, in NTv3, JEPA-DNA outperforms the baselines on 8 out of 9 supervised tasks, with less than 1\% degradation on the remaining task. In contrast, for HyenaDNA, although JEPA-DNA yields substantial improvements (e.g., a 9\%--11\% gain on the GUE promoter task) and overall positive effects (improvements in 6 out of 9 tasks), it also introduces notable degradations in certain cases, such as an 8\% decrease in performance on the GUE TF binding task.

Finally, we compare JEPA-DNA against seven leading GFMs evaluated in their original form. JEPA-DNA achieves state-of-the-art results under our linear probing settings on 7 out of 9 tasks. We indicate state-of-the-art performance with a '$\dagger$' symbol in Table 1 and provide the full per-model breakdown and full comparison in Tables~\ref{tab:baselines}--\ref{tab:jepa_vs_sota}, respectively.

\begin{table}[h!]
\centering
\scriptsize
\renewcommand{\arraystretch}{1.1}
\caption{Zero-shot performance of JEPA-DNA using the DNABERT-2 backbone (AUROC), compared against the original checkpoint and the two continually-trained native baselines. Gain denotes the \emph{relative} AUROC gain of JEPA-DNA over the original checkpoint / step-matched / compute-matched models. Bold values denote superior zero-shot performance.}
\label{tab:zs_dnabert}
{\color{black}
%\resizebox{\textwidth}{!}{
\begin{tabular}{@{}l cccc r@{}}
\toprule
\textbf{Task} & \textbf{Original} & \textbf{Step-} & \textbf{Compute-} & \textbf{JEPA-DNA} & \textbf{Gain} \\
              & \textbf{checkpoint} & \textbf{matched} & \textbf{matched} & \textbf{(Ours)} & \scriptsize \% AUROC \\
\midrule
BEND Expression Effect   & 0.543 & 0.530 & 0.536 & \textbf{0.574} & +5.71 / +8.21 / +7.00 \\
BEND Disease Variant     & 0.680 & 0.638 & 0.631 & \textbf{0.762} & +12.06 / +19.48 / +20.68 \\
TraitGym Complex         & \textbf{0.517} & 0.512 & 0.514 & 0.503 & -2.71 / -1.80 / -2.06 \\
TraitGym Mendelian       & 0.515 & 0.530 & 0.526 & \textbf{0.540} & +4.85 / +1.93 / +2.57 \\
Songlab ClinVar          & \textbf{0.540} & 0.518 & 0.516 & 0.527 & -2.41 / +1.72 / +2.15 \\
LOL-EVE Causal eQTL      & 0.515 & 0.522 & 0.523 & \textbf{0.545} & +5.83 / +4.47 / +4.26 \\
BRCA1                    & 0.505 & 0.522 & 0.501 & \textbf{0.527} & +4.36 / +0.93 / +5.12 \\
LRB Pathogenic OMIM      & 0.537 & 0.516 & 0.503 & \textbf{0.565} & +5.21 / +9.49 / +12.43 \\
\bottomrule
\end{tabular}
}%}
\end{table}

\subsection{Zero-Shot Evaluation}
We assess out-of-the-box semantic understanding by computing the cosine similarity between reference and variant sequence embeddings. We quantify performance via AUROC over the resulting similarity scores, allowing us to measure the model's capacity to rank functional variants without task-specific training.

Table~\ref{tab:zs_dnabert} summarizes these results  using DNABERT-2 as the representative backbone. Compared to the original checkpoint of DNABERT-2, JEPA-DNA improves zero-shot performance in 6 out of 8 tasks, with a mean AUROC gain of +4.1\% and substantial improvements (ranging from 4\% to 12\%) across clinically significant benchmarks tasks, including BEND Disease Variant (+12.1\%) and LRB Pathogenic OMIM (+5.21\%).

While JEPA-DNA provides broad benefits, the task-specific gain vary between architectures (additional backbones and metrics are reported in  Tables~\ref{tab:zs_continual_baselines}--~\ref{tab:zs_detailed}). In zero shot tasks, where the model did not receive direct supervision on functional or biological meaning, all evaluated models struggle, and performance remains low. However, JEPA-DNA recovers additional signal and improves performance in some cases, for example, in BEND Expression Effect and TraitGym Mendelian tasks.
Continual training alone yields little to no benefit, improving only 10–12 of 24 task-backbone pairs with average relative AUROC changes of $-0.60\%$ to $-0.21\%$ for step-matched and compute-matched setups, respectively. In contrast, JEPA-DNA outperforms these compute- and step-matched baselines on 17 of 24 pairs, achieving mean relative gains of $+3.65\%$ and $+3.31\%$, correspondingly.
% Whereas absolute zero-shot performance remains low for all models due to severe class imbalance, JEPA-DNA’s consistent ranking improvements on these tasks give an indicative signal that the pre-training objective has captured biologically relevant structure. 

\subsection{Ablations and Additional Experiments}

\begin{table}[tbh!]
\centering
\scriptsize
\setlength{\tabcolsep}{4pt}
\renewcommand{\arraystretch}{1.1}
\caption{Factorial ablation of the JEPA-DNA components on DNABERT-2. Promoter, Splice and Express denote GUE Promoter, GUE Splice Site and BEND Expression Effect, respectively. Each entry reports \emph{step-matched} (100K optimization steps) / \emph{compute-matched} (equal GPU-hours) results; Avg. Gain is the mean \emph{relative} AUROC gain over the baseline.}
\label{tab:ablation_jepa_dna}
{\color{black}
%\resizebox{\textwidth}{!}{%
\begin{tabular}{@{}l ccccc ccc r@{}}
\toprule
\textbf{Configuration} & \multicolumn{5}{c}{\textbf{Components}} & \multicolumn{3}{c}{\textbf{AUROC} \scriptsize (step / compute)} & \textbf{Avg. Gain} \\
\cmidrule(lr){2-6} \cmidrule(lr){7-9}
& \scriptsize MLM & \scriptsize EMA & \scriptsize JEPA & \scriptsize VICReg & \scriptsize Sched. Mask & \scriptsize Promoter & \scriptsize Splice & \scriptsize Express. & \scriptsize \% AUROC \\
\midrule
Baseline (DNABERT-2) & $\times$ & $\times$ & $\times$ & $\times$ & $\times$ & 0.884 & 0.653 & 0.543 & --- \\
\midrule
MLM & \checkmark & $\times$ & $\times$ & $\times$ & $\times$ & 0.927/0.927 & 0.670/0.672 & 0.530/0.536 & 1.71/2.16 \\
MLM+SM & \checkmark & $\times$ & $\times$ & $\times$ & \checkmark & 0.927/0.927 & 0.669/0.669 & 0.530/0.530 & 1.66/1.64 \\
MLM+EMA & \checkmark & \checkmark & $\times$ & $\times$ & $\times$ & 0.929/0.929 & 0.674/0.674 & 0.532/0.532 & 2.09/2.09 \\
MLM+SM+EMA & \checkmark & \checkmark & $\times$ & $\times$ & \checkmark & 0.927/0.927 & 0.669/0.669 & 0.530/0.530 & 1.67/1.64 \\
MLM+VR+SM \textit{(no JEPA)} & \checkmark & \checkmark & $\times$ & \checkmark & \checkmark & 0.924/0.925 & 0.678/0.679 & 0.547/0.549 & 3.06/3.24 \\
JEPA+VR+SM \textit{(no MLM)} & $\times$ & \checkmark & \checkmark & \checkmark & \checkmark & 0.919/0.920 & 0.680/0.676 & 0.562/0.565 & 3.86/3.85 \\
MLM+JEPA & \checkmark & \checkmark & \checkmark & $\times$ & $\times$ & 0.927/0.927 & 0.669/0.670 & 0.529/0.529 & 1.58/1.63 \\
MLM+JEPA+SM \textit{(no VR)} & \checkmark & \checkmark & \checkmark & $\times$ & \checkmark & 0.925/0.925 & 0.687/0.687 & 0.519/0.525 & 1.81/2.18 \\
MLM+JEPA+VR \textit{(no SM)} & \checkmark & \checkmark & \checkmark & \checkmark & $\times$ & 0.928/0.928 & 0.689/0.689 & 0.565/0.565 & 4.85/4.85 \\
\midrule
\textbf{JEPA-DNA (full)} & \checkmark & \checkmark & \checkmark & \checkmark & \checkmark & \textbf{0.928} & \textbf{0.705} & \textbf{0.574} & \textbf{6.22/6.22} \\
\bottomrule
\end{tabular}}
%}
\end{table}

%\rev{Through extensive compute- and step- matched ablations on the training objectives, predictor architectures, and optimization strategies, we establish the main drivers of performance gains, confirming that improvements stem from the main components of JEPA-DNA rather than merely continual training.}
\textbf{Ablations.} We systematically decompose the JEPA-DNA framework to identify the primary drivers of performance gains, using DNABERT-2 as the backbone and evaluating it on three representative tasks. Table~\ref{tab:ablation_jepa_dna} details the incremental utility of our framework's components, where VICReg (VR) regularization emerges as essential for unlocking the benefits of the hybrid MLM+JEPA paradigm. Peak performance is achieved through the synergistic combination of regularized latent prediction and Scheduled Masking (SM). The grid is reported under both a step-matched and a compute-matched budget and includes a single-component elimination control for each part of the framework.

We also compare latent-space prediction against reconstruction-based paradigms for continual training. Table~\ref{tab:ablation_training_obj} in the Appendix shows that latent-space alignment (+3.86\%) provides a more informative training signal than token reconstruction (+2.06\%). Optimal results require a hybrid approach that leverages MLM tokens and local reconstruction alongside latent sequence alignment.

We further evaluate our architectural and optimization choices. Success in latent alignment depends on the predictor architecture. As shown in Table~\ref{tab:predictor_arch} in the Appendix, an attention-based predictor substantially outperforms the MLP variant, with performance peaking at a depth of four layers. We also adopt the target encoder, consistent with prior JEPA approaches, and employ it for inference. We also find that the SGDM algorithm is essential for training stability, as standard adaptive optimizers such as Adam or AdamW frequently lead to performance degradation (see Appendix Table~\ref{tab:optimizers}). The extended ablation analysis is provided in Appendix~\ref{subsec:appendix_ablation}.

\textbf{Additional Experiments.} We briefly summarize five additional experiments. Compared against DNASimCLR~\citep{yang2024dnasimclr} and three matched contrastive variants, JEPA-DNA is ahead on all evaluated tasks ($+7.57\%$ against $+4.06\%$ for the strongest variant, Table~\ref{tab:dnasimclr} in the Appendix). Additionally, Appendix~\ref{subsec:appendix_results} details an embedding analysis (Figure~\ref{fig:umap}), a full fine-tuning comparison (Table~\ref{tab:finetune}), a sequence-length stability analysis (Table~\ref{tab:length_sweep}) and a loss-weight sensitivity analysis (Table~\ref{tab:loss_weights}).

\section{Conclusion}\label{sec:conclusion}

\textbf{Summary.} We introduced JEPA-DNA, a framework adapting the Joint Embedding Predictive Architecture to genomic sequences. By predicting latent representations in addition to reconstructing tokens, JEPA-DNA captures high-level biological context more effectively than generative objectives alone, consistently boosting backbone performance across supervised and zero-shot tasks.

\textbf{Limitations and Future Work. } While our work focused on representation quality, the behaviour of full fine-tuning remains partially addressed, as our end-to-end comparison  covers only one backbone and a subset of tasks (Appendix Table~\ref{tab:finetune}). A broader study spanning full fine-tuning and parameter-efficient adaptation is deferred for future work. In addition, due to the substantial computational demands of large-scale genomic training, this study focused on the continual pre-training of existing state-of-the-art backbones. Future work will evaluate the JEPA objective for training models from scratch on massive, multi-species datasets to determine if it accelerates the emergence of biological priors. Furthermore, integrating recent advancements such as LeJEPA~\citep{balestriero2025lejepa}, could potentially offer better scaling properties for the extremely long-range dependencies inherent in genomic data.

\bibliographystyle{unsrt}
\bibliography{gfm}

@article{ji2021dnabert,
  title={DNABERT: pre-trained Bidirectional Encoder Representations from Transformers model for DNA-language in genome},
  author={Ji, Yanrong and Zhou, Zhihan and Liu, Han and Davuluri, Ramana V},
  journal={Bioinformatics},
  volume={37},
  number={15},
  pages={2112--2120},
  year={2021},
  publisher={Oxford University Press}
}

@article{nguyen2024sequence,
  title={Sequence modeling and design from molecular to genome scale with Evo},
  author={Nguyen, Eric and Poli, Michael and Durrant, Matthew G and Kang, Brian and Katrekar, Dhruva and Li, David B and Bartie, Liam J and Thomas, Armin W and King, Samuel H and Brixi, Garyk and others},
  journal={Science},
  volume={386},
  number={6723},
  pages={eado9336},
  year={2024},
  publisher={American Association for the Advancement of Science}
}

@inproceedings{zhou2024dnabert,
  title={DNABERT-2: Efficient foundation model and benchmark for multi-species genomes},
  author={Zhou, Zhihan and Ji, Yanrong and Li, Weijian and Dutta, Pratik and Davuluri, Ramana and Liu, Han},
  booktitle={International Conference on Learning Representations},
  volume={2024},
  pages={41642--41665},
  year={2024}
}

@article{dalla2025nucleotide,
  title={Nucleotide transformer: building and evaluating robust foundation models for human genomics},
  author={Dalla-Torre, Hugo and Gonzalez, Liam and Mendoza-Revilla, Javier and Lopez Carranza, Nicolas and Grzywaczewski, Adam Henryk and Oteri, Francesco and Dallago, Christian and Trop, Evan and De Almeida, Bernardo P and Sirelkhatim, Hassan and others},
  journal={Nature methods},
  volume={22},
  number={2},
  pages={287--297},
  year={2025},
  publisher={Nature Publishing Group US New York}
}

@article{boshar2025foundational,
  title={A foundational model for joint sequence-function multi-species modeling at scale for long-range genomic prediction},
  author={Boshar, Sam and Evans, Benjamin and Tang, Ziqi and Picard, Armand and Adel, Yanis and Lorbeer, Franziska K and Rajesh, Chandana and Karch, Tristan and Sidbon, Shawn and Emms, David and others},
  journal={BioRxiv},
  pages={2025--12},
  year={2025},
  publisher={Cold Spring Harbor Laboratory}
}

@article{schiff2024caduceus,
  title={Caduceus: Bi-directional equivariant long-range dna sequence modeling},
  author={Schiff, Yair and Kao, Chia-Hsiang and Gokaslan, Aaron and Dao, Tri and Gu, Albert and Kuleshov, Volodymyr},
  journal={Proceedings of machine learning research},
  volume={235},
  pages={43632},
  year={2024}
}

@article{nguyen2023hyenadna,
  title={Hyenadna: Long-range genomic sequence modeling at single nucleotide resolution},
  author={Nguyen, Eric and Poli, Michael and Faizi, Marjan and Thomas, Armin and Wornow, Michael and Birch-Sykes, Callum and Massaroli, Stefano and Patel, Aman and Rabideau, Clayton and Bengio, Yoshua and others},
  journal={Advances in neural information processing systems},
  volume={36},
  pages={43177--43201},
  year={2023}
}

@article{brixi2026genome,
  title={Genome modelling and design across all domains of life with Evo 2},
  author={Brixi, Garyk and Durrant, Matthew G and Ku, Jerome and Naghipourfar, Mohsen and Poli, Michael and Sun, Gwanggyu and Brockman, Greg and Chang, Daniel and Fanton, Alison and Gonzalez, Gabriel A and others},
  journal={Nature},
  volume={652},
  number={8112},
  pages={1349--1361},
  year={2026},
  publisher={Nature Publishing Group UK London}
}

@article{benegas2025dna,
  title={A DNA language model based on multispecies alignment predicts the effects of genome-wide variants},
  author={Benegas, Gonzalo and Albors, Carlos and Aw, Alan J and Ye, Chengzhong and Song, Yun S},
  journal={Nature Biotechnology},
  volume={43},
  number={12},
  pages={1960--1965},
  year={2025},
  publisher={Nature Publishing Group US New York}
}

@inproceedings{marin2024bend,
  title={Bend: Benchmarking dna language models on biologically meaningful tasks},
  author={Marin, Frederikke and Teufel, Felix and Horlacher, Marc and Madsen, Dennis and Pultz, Dennis and Winther, Ole and Boomsma, Wouter},
  booktitle={International Conference on Learning Representations},
  volume={2024},
  pages={15246--15281},
  year={2024}
}

@article{medvedev2025biotoken,
  title={BioToken and BioFM--biologically-informed tokenization enables accurate and efficient genomic foundation models},
  author={Medvedev, Aleksandr and Viswanathan, Karthik and Kanithi, Praveenkumar and Vishniakov, Kirill and Munjal, Prateek and Christophe, Clement and Magalhães, Tiago R and Pimentel, Marco AF and Rajan, Ronnie and Khan, Shadab},
  journal={bioRxiv},
  pages={2025--03},
  year={2025},
  publisher={Cold Spring Harbor Laboratory}
}

@inproceedings{kao2024advancing,
  title={Advancing dna language models: The genomics long-range benchmark},
  author={Kao, Chia Hsiang and Trop, Evan and Polen, McKinley and Schiff, Yair and de Almeida, Bernardo P and Gokaslan, Aaron and Pierrot, Thomas and Kuleshov, Volodymyr},
  booktitle={ICLR 2024 Workshop on Machine Learning for Genomics Explorations},
  year={2024}
}

@article{benegas2025benchmarking,
  title={Benchmarking dna sequence models for causal regulatory variant prediction in human genetics},
  author={Benegas, Gonzalo and Eraslan, G{\"o}kcen and Song, Yun S},
  journal={bioRxiv},
  year={2025}
}

@article{findlay2018accurate,
  title={Accurate classification of BRCA1 variants with saturation genome editing},
  author={Findlay, Gregory M and Daza, Riza M and Martin, Beth and Zhang, Melissa D and Leith, Anh P and Gasperini, Molly and Janizek, Joseph D and Huang, Xingfan and Starita, Lea M and Shendure, Jay},
  journal={Nature},
  volume={562},
  number={7726},
  pages={217--222},
  year={2018},
  publisher={Nature Publishing Group UK London}
}

@article{shearer2024genomic,
  title={A Genomic Language Model for Zero-Shot Prediction of Promoter Variant Effects},
  author={Shearer, Courtney A and Orenbuch, Rose and Teufel, Felix and Steinmetz, Christian J and Ritter, Daniel and Xie, Erik and Gazizov, Artem and Spinner, Aviv and Frazer, Jonathan and Dias, Mafalda and others},
  journal={bioRxiv},
  pages={2024--11},
  year={2024},
  publisher={Cold Spring Harbor Laboratory}
}

@article{schneider2017evaluation,
  title={Evaluation of GRCh38 and de novo haploid genome assemblies demonstrates the enduring quality of the reference assembly},
  author={Schneider, Valerie A and Graves-Lindsay, Tina and Howe, Kerstin and Bouk, Nathan and Chen, Hsiu-Chuan and Kitts, Paul A and Murphy, Terence D and Pruitt, Kim D and Thibaud-Nissen, Fran{\c{c}}oise and Albracht, Derek and others},
  journal={Genome research},
  volume={27},
  number={5},
  pages={849--864},
  year={2017},
  publisher={Cold Spring Harbor Laboratory Press}
}

@article{benegas2025genomic,
  title={Genomic language models: opportunities and challenges},
  author={Benegas, Gonzalo and Ye, Chengzhong and Albors, Carlos and Li, Jianan Canal and Song, Yun S},
  journal={Trends in Genetics},
  volume={41},
  number={4},
  pages={286--302},
  year={2025},
  publisher={Elsevier}
}

@article{lecun2022path,
  title={A path towards autonomous machine intelligence version 0.9. 2, 2022-06-27},
  author={LeCun, Yann and others},
  journal={Open Review},
  volume={62},
  number={1},
  pages={1--62},
  year={2022}
}

@article{larey2026gfmbench,
  title={GFMBench-API: A Standardized Interface for Benchmarking Genomic Foundation Models},
  author={Larey, Ariel and Dahan, Elay and Bleiweiss, Amit and Kellerman, Raizy and Leib, Guy and Nayshool, Omri and Ofer, Dan and Zinger, Tal and Dominissini, Dan and Rechavi, Gideon and others},
  journal={Biorxiv},
  pages={2026--02},
  year={2026},
  publisher={Cold Spring Harbor Laboratory}
}

@inproceedings{devlin2019bert,
  title={Bert: Pre-training of deep bidirectional transformers for language understanding},
  author={Devlin, Jacob and Chang, Ming-Wei and Lee, Kenton and Toutanova, Kristina},
  booktitle={Proceedings of the 2019 conference of the North American chapter of the association for computational linguistics: human language technologies, volume 1 (long and short papers)},
  pages={4171--4186},
  year={2019}
}

@article{benegas2023dna,
  title={DNA language models are powerful predictors of genome-wide variant effects},
  author={Benegas, Gonzalo and Batra, Sanjit Singh and Song, Yun S},
  journal={Proceedings of the National Academy of Sciences},
  volume={120},
  number={44},
  pages={e2311219120},
  year={2023},
  publisher={National Academy of Sciences}
}

@article{yang2024dnasimclr,
  title={DNASimCLR: a contrastive learning-based deep learning approach for gene sequence data classification},
  author={Yang, Minghao and Wang, Zehua and Yan, Zizhuo and Wang, Wenxiang and Zhu, Qian and Jin, Changlong},
  journal={BMC bioinformatics},
  volume={25},
  number={1},
  pages={328},
  year={2024},
  publisher={Springer}
}

@inproceedings{assran2023self,
  title={Self-supervised learning from images with a joint-embedding predictive architecture},
  author={Assran, Mahmoud and Duval, Quentin and Misra, Ishan and Bojanowski, Piotr and Vincent, Pascal and Rabbat, Michael and LeCun, Yann and Ballas, Nicolas},
  booktitle={2023 IEEE/CVF Conference on Computer Vision and Pattern Recognition (CVPR)},
  pages={15619--15629},
  year={2023},
  organization={IEEE}
}

@inproceedings{huang2026llm,
  title={Llm-jepa: Large language models meet joint embedding predictive architectures},
  author={Huang, Hai and LeCun, Yann and Balestriero, Randall},
  booktitle={International Conference on Learning Representations},
  volume={2026},
  pages={105717--105737},
  year={2026}
}

@article{litman2025genejepa,
  title={Genejepa: A predictive world model of the transcriptome},
  author={Litman, Elon and Myers, Tyler and Agarwal, Vinayak and Mittal, Ekansh and Li, Orion and Gopinath, Ashwin and Kassis, Timothy},
  journal={bioRxiv},
  pages={2025--10},
  year={2025},
  publisher={Cold Spring Harbor Laboratory}
}

@article{church2011modernizing,
  title={Modernizing reference genome assemblies},
  author={Church, Deanna M and Schneider, Valerie A and Graves, Tina and Auger, Katherine and Cunningham, Fiona and Bouk, Nathan and Chen, Hsiu-Chuan and Agarwala, Richa and McLaren, William M and Ritchie, Graham RS and others},
  journal={PLoS biology},
  volume={9},
  number={7},
  pages={e1001091},
  year={2011},
  publisher={Public Library of Science San Francisco, USA}
}

@article{howe2013zebrafish,
  title={The zebrafish reference genome sequence and its relationship to the human genome},
  author={Howe, Kerstin and Clark, Matthew D and Torroja, Carlos F and Torrance, James and Berthelot, Camille and Muffato, Matthieu and Collins, John E and Humphray, Sean and McLaren, Karen and Matthews, Lucy and others},
  journal={Nature},
  volume={496},
  number={7446},
  pages={498--503},
  year={2013},
  publisher={Nature Publishing Group UK London}
}

@article{hoskins2015release,
  title={The Release 6 reference sequence of the Drosophila melanogaster genome},
  author={Hoskins, Roger A and Carlson, Joseph W and Wan, Kenneth H and Park, Soo and Mendez, Ivonne and Galle, Samuel E and Booth, Benjamin W and Pfeiffer, Barret D and George, Reed A and Svirskas, Robert and others},
  journal={Genome research},
  volume={25},
  number={3},
  pages={445--458},
  year={2015},
  publisher={Cold Spring Harbor Laboratory Press}
}

@article{c1998genome,
  title={Genome sequence of the nematode C. elegans: a platform for investigating biology},
  author={C. elegans Sequencing Consortium*},
  journal={Science},
  volume={282},
  number={5396},
  pages={2012--2018},
  year={1998},
  publisher={American Association for the Advancement of Science}
}

@article{arabidopsis2000analysis,
  title={Analysis of the genome sequence of the flowering plant Arabidopsis thaliana},
  author={Arabidopsis Genome Initiative genomeanalysis@ tgr. org genomeanalysis@ gsf. de},
  journal={nature},
  volume={408},
  number={6814},
  pages={796--815},
  year={2000},
  publisher={Nature Publishing Group UK London}
}

@article{kent2002human,
  title={The human genome browser at UCSC},
  author={Kent, W James and Sugnet, Charles W and Furey, Terrence S and Roskin, Krishna M and Pringle, Tom H and Zahler, Alan M and Haussler, David},
  journal={Genome research},
  volume={12},
  number={6},
  pages={996--1006},
  year={2002},
  publisher={Cold Spring Harbor Laboratory Press}
}

@article{bardes2021vicreg,
  title={Vicreg: Variance-invariance-covariance regularization for self-supervised learning},
  author={Bardes, Adrien and Ponce, Jean and LeCun, Yann},
  journal={arXiv preprint arXiv:2105.04906},
  year={2021}
}

@article{banovich2014methylation,
  title={Methylation QTLs are associated with coordinated changes in transcription factor binding, histone modifications, and gene expression levels},
  author={Banovich, Nicholas E and Lan, Xun and McVicker, Graham and Van de Geijn, Bryce and Degner, Jacob F and Blischak, John D and Roux, Julien and Pritchard, Jonathan K and Gilad, Yoav},
  journal={PLoS genetics},
  volume={10},
  number={9},
  pages={e1004663},
  year={2014},
  publisher={Public Library of Science San Francisco, USA}
}

@article{wang2013transcription,
  title={Transcription factor and chromatin features predict genes associated with eQTLs},
  author={Wang, Dennis and Rendon, Augusto and Wernisch, Lorenz},
  journal={Nucleic acids research},
  volume={41},
  number={3},
  pages={1450--1463},
  year={2013},
  publisher={Oxford University Press}
}

@article{zeng2022predicting,
  title={Predicting RNA splicing from DNA sequence using Pangolin},
  author={Zeng, Tony and Li, Yang I},
  journal={Genome biology},
  volume={23},
  number={1},
  pages={103},
  year={2022},
  publisher={Springer}
}

@article{balestriero2025lejepa,
  title={Lejepa: Provable and scalable self-supervised learning without the heuristics},
  author={Balestriero, Randall and LeCun, Yann},
  journal={arXiv preprint arXiv:2511.08544},
  year={2025}
}

@inproceedings{chen2020simple,
  title={A simple framework for contrastive learning of visual representations},
  author={Chen, Ting and Kornblith, Simon and Norouzi, Mohammad and Hinton, Geoffrey},
  booktitle={International conference on machine learning},
  pages={1597--1607},
  year={2020},
  organization={PmLR}
}

@article{harrison2024ensembl,
  title={Ensembl 2024},
  author={Harrison, Peter W and Amode, M Ridwan and Austine-Orimoloye, Olanrewaju and Azov, Andrey G and Barba, Matthieu and Barnes, If and Becker, Arne and Bennett, Ruth and Berry, Andrew and Bhai, Jyothish and others},
  journal={Nucleic acids research},
  volume={52},
  number={D1},
  pages={D891--D899},
  year={2024},
  publisher={Oxford University Press}
}

@article{kingma2014adam,
  title={Adam: A method for stochastic optimization},
  author={Kingma, Diederik P and Ba, Jimmy},
  journal={arXiv preprint arXiv:1412.6980},
  year={2014}
}

@article{loshchilov2017decoupled,
  title={Decoupled weight decay regularization},
  author={Loshchilov, Ilya and Hutter, Frank},
  journal={arXiv preprint arXiv:1711.05101},
  year={2017}
}

@article{zhuang2020adabelief,
  title={Adabelief optimizer: Adapting stepsizes by the belief in observed gradients},
  author={Zhuang, Juntang and Tang, Tommy and Ding, Yifan and Tatikonda, Sekhar C and Dvornek, Nicha and Papademetris, Xenophon and Duncan, James},
  journal={Advances in neural information processing systems},
  volume={33},
  pages={18795--18806},
  year={2020}
}

@inproceedings{bottou2010large,
  title={Large-scale machine learning with stochastic gradient descent},
  author={Bottou, L{\'e}on},
  booktitle={Proceedings of COMPSTAT'2010: 19th International Conference on Computational Statistics, Paris France, August 22-27, 2010 Keynote, Invited and Contributed Papers},
  pages={177--186},
  year={2010},
  organization={Springer}
}

@article{cheng2025dnalongbench,
  title={DNALONGBENCH: a benchmark suite for long-range DNA prediction tasks},
  author={Cheng, Wenduo and Song, Zhenqiao and Zhang, Yang and Wang, Shike and Wang, Danqing and Yang, Muyu and Li, Lei and Ma, Jian},
  journal={Nature communications},
  volume={16},
  number={1},
  pages={10108},
  year={2025},
  publisher={Nature Publishing Group UK London}
}

%%%%%%%%%%%%%%%%%%%%%%%%%%%%%%%%%%%%%%%%%%%%%%%%%%%%%%%%%%%%
\clearpage
\appendix
\section{Appendix}
This appendix provides supplementary information to support the findings and methodology presented in the main text. 

Section~\ref{subsec:appendix_cost} quantifies the training and inference cost of the framework.

Section~\ref{subsec:appendix_results} offers extended experimental results, including robustness (multi-seed) analysis, detailed performance breakdowns, and additional ablations, as well as comparisons against continually-trained baselines, a full fine-tuning study, loss-weight sensitivity, and an analysis of behaviour beyond the training context.

Section~\ref{subsec:appendix_impl_details} provides a comprehensive description of the experimental setup, covering backbone and architecture specifications, dataset and task descriptions, evaluation protocols, and specific implementation and training details required for reproducibility.

\subsection{Computational Cost}\label{subsec:appendix_cost}
Table~\ref{tab:cost} reports the training cost of JEPA-DNA relative to matched native continual training. The overhead stems from the additional forward passes through the EMA target encoder and the predictor, and ranges from $1.09\times$ to $7.88\times$ in time per step depending on the backbone. This overhead is confined entirely to the continual pre-training phase. At inference the predictor and the VICReg heads are discarded and only the EMA target encoder is retained, so runtime and memory are identical to those of the original backbone.

\begin{table}[tbh!]
\centering
\scriptsize
\renewcommand{\arraystretch}{1.1}
\caption{Computational cost of JEPA-DNA relative to matched native continual training. The overhead is confined to pre-training; inference cost is unchanged.}
\label{tab:cost}
{\color{black}
\begin{tabular}{@{}ll ccc@{}}
\toprule
\textbf{Backbone} & \textbf{Cost} & \textbf{Native MLM/NTP} & \textbf{JEPA-DNA} & \textbf{Ratio} \\
\midrule
DNABERT-2 & Time / step & 0.955 s & 2.661 s & 2.78$\times$ \\
          & Peak memory & 21.34 GiB & 49.42 GiB & 2.32$\times$ \\
          & GPU-hours (100K steps) & $\sim$46.3 h & $\sim$128.9 h & 2.78$\times$ \\
\midrule
NTv3-100M & Time / step & 4.654 s & 5.055 s & 1.09$\times$ \\
          & Peak memory & 37.45 GiB & 71.82 GiB & 1.92$\times$ \\
          & GPU-hours (32.5K steps) & $\sim$40.2 h & $\sim$43.6 h & 1.09$\times$ \\
\midrule
HyenaDNA  & Time / step & 0.631 s & 4.966 s & 7.88$\times$ \\
          & Peak memory & 3.80 GiB & 15.11 GiB & 3.98$\times$ \\
          & GPU-hours (45K steps) & $\sim$14.0 h & $\sim$110.0 h & 7.88$\times$ \\
\midrule
\textit{All} & \textbf{Inference time / memory} & --- & --- & \textbf{1.00$\times$} \\
\bottomrule
\end{tabular}
}
\end{table}

\begin{table}[tbh!]
\centering
\scriptsize
\renewcommand{\arraystretch}{1.1}
\caption{Linear probing performance across diverse GFM backbones, objectives, aggregation strategies, and self-supervised datasets (DS). Bold values indicate the top performance within each backbone group. Gain denotes the \emph{relative} AUROC gain over the corresponding original checkpoint baseline.}
\label{tab:supervised_detailed}
\begin{tabular}{@{}l ccc ccc r@{}}
\toprule
\textbf{Task} & \multicolumn{3}{c}{\textbf{Baseline}} & \multicolumn{3}{c}{\textbf{JEPA-DNA (Ours)}} & \textbf{Gain} \\
\cmidrule(lr){2-4} \cmidrule(lr){5-7} \cmidrule(lr){8-8}
& \scriptsize MCC & \scriptsize AUROC & \scriptsize AUPRC & \scriptsize MCC & \scriptsize AUROC & \scriptsize AUPRC & \scriptsize \% AUROC \\
\midrule
\multicolumn{8}{l}{\textit{DNABERT-2 (Transformer + MLM + CLS; Multi-Species DS)}} \\
\midrule
GUE TF Binding    & 0.429 & 0.788 & 0.773 & \textbf{0.515} & \textbf{0.839} & \textbf{0.828} & +6.47\% \\
GUE Promoter      & 0.607 & 0.884 & 0.877 & \textbf{0.717} & \textbf{0.928} & \textbf{0.913} & +4.98\% \\
GUE Splice Site   & 0.000 & 0.653 & 0.452 & \textbf{0.139} & \textbf{0.705} & \textbf{0.500} & +7.96\% \\
VB Coding Path.   & 0.002 & \textbf{0.624} & \textbf{0.513} & \textbf{0.078} & 0.610 & 0.509 & -2.24\% \\
% VB Non-coding     & 0.000 & 0.605 & 0.149 & 0.000 & \textbf{0.612} & \textbf{0.150} & +1.16\% \\
VB Exp. Effect    & 0.254 & 0.674 & 0.647 & \textbf{0.273} & \textbf{0.689} & \textbf{0.658} & +2.23\% \\
VB Common/Rare    & 0.007 & \textbf{0.505} & \textbf{0.504} & 0.007 & 0.504 & 0.503 & -0.20\% \\
VB meQTL          & 0.032 & 0.623 & 0.552 & \textbf{0.125} & \textbf{0.645} & \textbf{0.579} & +3.53\% \\
VB sQTL           & \textbf{0.121} & \textbf{0.585} & \textbf{0.587} & 0.110 & 0.577 & 0.582 & -1.37\% \\
LRB Causal eQTL   & \textbf{0.307} & 0.697 & 0.717 & 0.296 & \textbf{0.711} & \textbf{0.727} & +2.01\% \\
\midrule
\multicolumn{8}{l}{\textit{Nucleotide Transformer v3 (Hybrid + MLM + Mean Pooling; Open-Genome 2 DS)}} \\
\midrule
GUE TF Binding    & 0.509 & 0.837 & 0.825 & \textbf{0.530} & \textbf{0.849} & \textbf{0.843} & +1.43\% \\
GUE Promoter      & 0.648 & 0.907 & 0.894 & \textbf{0.754} & \textbf{0.937} & \textbf{0.923} & +3.31\% \\
GUE Splice Site   & -0.004 & 0.713 & 0.487 & \textbf{0.163} & \textbf{0.733} & \textbf{0.506} & +2.81\% \\
VB Coding Path.   & 0.017 & 0.668 & 0.541 & \textbf{0.099} & \textbf{0.675} & \textbf{0.554} & +1.05\% \\
% VB Non-coding     & 0.000 & 0.625 & 0.156 & 0.000 & \textbf{0.650} & \textbf{0.237} & +4.00\% \\
VB Exp. Effect    & 0.223 & 0.642 & 0.612 & \textbf{0.235} & \textbf{0.654} & \textbf{0.626} & +1.87\% \\
VB Common/Rare    & \textbf{0.003} & 0.502 & 0.503 & -0.001 & \textbf{0.510} & \textbf{0.508} & +1.59\% \\
VB meQTL          & -0.008 & 0.576 & 0.496 & \textbf{0.039} & \textbf{0.615} & \textbf{0.545} & +6.77\% \\
VB sQTL           & 0.131 & \textbf{0.591} & 0.591 & \textbf{0.135} & 0.590 & 0.591 & -0.17\% \\
LRB Causal eQTL   & 0.269 & 0.690 & 0.675 & \textbf{0.289} & \textbf{0.697} & \textbf{0.686} & +1.01\% \\
\midrule
\multicolumn{8}{l}{\textit{HyenaDNA (Long Convolutions + NTP + last-token; Reference-Genome DS)}} \\
\midrule
GUE TF Binding    & \textbf{0.278} & \textbf{0.698} & \textbf{0.684} & 0.084 & 0.638 & 0.608 & -8.60\% \\
GUE Promoter      & 0.296 & 0.686 & 0.721 & \textbf{0.434} & \textbf{0.763} & \textbf{0.728} & +11.22\% \\
GUE Splice Site   & -0.012 & 0.506 & 0.336 & \textbf{0.000} & \textbf{0.558} & \textbf{0.375} & +10.28\% \\
VB Coding Path.   & -0.001 & 0.515 & 0.444 & \textbf{0.019} & \textbf{0.559} & \textbf{0.486} & +8.54\% \\
% VB Non-coding     & 0.000 & 0.537 & 0.116 & 0.000 & \textbf{0.582} & \textbf{0.137} & +8.38\% \\
VB Exp. Effect    & \textbf{0.110} & \textbf{0.580} & \textbf{0.563} & 0.108 & 0.547 & 0.523 & -5.69\% \\
VB Common/Rare    & 0.010 & 0.503 & 0.500 & \textbf{0.016} & \textbf{0.504} & \textbf{0.505} & +0.20\% \\
VB meQTL          & \textbf{0.021} & 0.503 & 0.461 & 0.020 & \textbf{0.518} & \textbf{0.471} & +2.98\% \\
VB sQTL           & 0.085 & 0.550 & \textbf{0.556} & \textbf{0.089} & \textbf{0.552} & 0.556 & +0.36\% \\
LRB Causal eQTL   & 0.108 & 0.565 & 0.567 & \textbf{0.189} & \textbf{0.623} & \textbf{0.615} & +10.27\% \\
\bottomrule
\end{tabular}
\end{table}

\subsection{Extended Results}\label{subsec:appendix_results}
\paragraph{Complete Supervised Metrics.}\label{subsec:appendix_full_metrics}
Table~\ref{tab:supervised_detailed} reports the full set of metrics for the original checkpoint and JEPA-DNA, namely the Matthews Correlation Coefficient (MCC), AUROC and AUPRC, across all backbones and supervised tasks.

\paragraph{Robustness Analysis.}\label{subsec:appendix_robustness}
We conduct a multi-seed analysis across all three backbones to evaluate the stability of the reported performance gains. Specifically, we repeat the linear-probe training over five random seeds for each backbone and report the mean and standard deviation. As shown in Table~\ref{tab:supervised_full_seeds}, JEPA-DNA yields consistent and stable improvements. For DNABERT-2, JEPA-DNA improves on 6 of 9 tasks, with gains of up to $+7.81\%$ on splice site prediction, and regresses on coding pathogenicity, common vs.\ rare and sQTL. For NTv3, JEPA-DNA improves on 8 of 9 tasks, the single exception being sQTL at $-0.08\%$, which lies inside one standard deviation. For HyenaDNA, JEPA-DNA improves on 6 of 9 tasks and regresses on TF binding, expression effect and common vs.\ rare. We report these degradations explicitly as trade-offs of the smallest backbone on locally-driven motif tasks. Where gains occur they are large relative to seed noise. For example, the NTv3 promoter improvement of $0.029$ absolute AUROC is roughly seventy times the corresponding standard deviation. Multi-seed results therefore preserve the trends observed in our single-run experiments.

\paragraph{Benchmarking Against Leading GFMs.}\label{subsec:appendix_sota}
To provide a comprehensive comparison against leading GFMs, Table~\ref{tab:baselines} details the per-task linear probing performance of each individual baseline GFM, reported as reference points for frozen representation quality. Table~\ref{tab:jepa_vs_sota} summarizes this comparison:

\begin{table}[tbh!]
\centering
\scriptsize
\renewcommand{\arraystretch}{1.2}
\caption{{Impact of JEPA-DNA on Representation Stability and Performance.} We compare each baseline backbone against its JEPA-DNA enhanced counterpart using linear probing across five random seeds, for all three backbones. We report the mean AUROC and AUPRC with standard deviations. Bold values indicate the superior mean performance. Gain denotes the \emph{relative} AUROC gain over the corresponding original checkpoint baseline.}
\label{tab:supervised_full_seeds}
\begin{tabular}{@{}l cc cc c@{}}
\toprule
\textbf{Task} & \multicolumn{2}{c}{\textbf{Baseline}} & \multicolumn{2}{c}{\textbf{JEPA-DNA (Ours)}} & \textbf{Gain} \\
\cmidrule(lr){2-3} \cmidrule(lr){4-5} \cmidrule(lr){6-6}
& \scriptsize AUROC & \scriptsize AUPRC & \scriptsize AUROC & \scriptsize AUPRC & \scriptsize \% AUROC \\
\midrule
\multicolumn{6}{l}{\textit{DNABERT-2 (Transformer + MLM + CLS; Multi-Species DS)}} \\
\midrule
GUE TF Binding & 0.787 $\pm$ .001 & 0.773 $\pm$ .001 & \textbf{0.839} $\pm$ .001 & \textbf{0.828} $\pm$ .001 & +6.61\% \\
GUE Promoter & 0.885 $\pm$ .001 & 0.877 $\pm$ .001 & \textbf{0.928} $\pm$ .000 & \textbf{0.913} $\pm$ .000 & +4.86\% \\
GUE Splice Site & 0.653 $\pm$ .001 & 0.451 $\pm$ .002 & \textbf{0.704} $\pm$ .002 & \textbf{0.500} $\pm$ .003 & +7.81\% \\
VB Coding Path. & \textbf{0.622} $\pm$ .003 & \textbf{0.511} $\pm$ .003 & 0.610 $\pm$ .005 & 0.511 $\pm$ .006 & -1.93\% \\
VB Exp. Effect & 0.673 $\pm$ .002 & 0.646 $\pm$ .001 & \textbf{0.688} $\pm$ .001 & \textbf{0.656} $\pm$ .001 & +2.23\% \\
VB Common/Rare & \textbf{0.508} $\pm$ .004 & \textbf{0.504} $\pm$ .006 & 0.505 $\pm$ .004 & 0.505 $\pm$ .003 & -0.59\% \\
VB meQTL & 0.619 $\pm$ .005 & 0.551 $\pm$ .004 & \textbf{0.645} $\pm$ .004 & \textbf{0.581} $\pm$ .005 & +4.20\% \\
VB sQTL & \textbf{0.585} $\pm$ .001 & \textbf{0.588} $\pm$ .001 & 0.579 $\pm$ .002 & 0.584 $\pm$ .002 & -1.03\% \\
LRB Causal eQTL & 0.696 $\pm$ .001 & 0.717 $\pm$ .001 & \textbf{0.708} $\pm$ .002 & \textbf{0.726} $\pm$ .001 & +1.72\% \\
\midrule
\multicolumn{6}{l}{\textit{Nucleotide Transformer v3 (Hybrid + MLM + Mean Pooling; Open-Genome 2 DS)}} \\
\midrule
GUE TF Binding & 0.837 $\pm$ .001 & 0.824 $\pm$ .001 & \textbf{0.848} $\pm$ .001 & \textbf{0.842} $\pm$ .001 & +1.33\% \\
GUE Promoter & 0.907 $\pm$ .000 & 0.894 $\pm$ .000 & \textbf{0.937} $\pm$ .000 & \textbf{0.923} $\pm$ .000 & +3.23\% \\
GUE Splice Site & 0.714 $\pm$ .002 & 0.488 $\pm$ .002 & \textbf{0.733} $\pm$ .002 & \textbf{0.506} $\pm$ .001 & +2.71\% \\
VB Coding Path. & 0.665 $\pm$ .002 & 0.539 $\pm$ .002 & \textbf{0.676} $\pm$ .002 & \textbf{0.555} $\pm$ .003 & +1.63\% \\
VB Exp. Effect & 0.642 $\pm$ .001 & 0.613 $\pm$ .001 & \textbf{0.653} $\pm$ .004 & \textbf{0.624} $\pm$ .005 & +1.68\% \\
VB Common/Rare & 0.505 $\pm$ .003 & 0.505 $\pm$ .003 & \textbf{0.513} $\pm$ .005 & \textbf{0.511} $\pm$ .004 & +1.76\% \\
VB meQTL & 0.578 $\pm$ .001 & 0.497 $\pm$ .001 & \textbf{0.614} $\pm$ .006 & \textbf{0.546} $\pm$ .008 & +6.39\% \\
VB sQTL & \textbf{0.591} $\pm$ .001 & \textbf{0.591} $\pm$ .001 & 0.590 $\pm$ .002 & 0.592 $\pm$ .001 & -0.08\% \\
LRB Causal eQTL & 0.689 $\pm$ .001 & 0.675 $\pm$ .000 & \textbf{0.697} $\pm$ .001 & \textbf{0.686} $\pm$ .001 & +1.13\% \\
\midrule
\multicolumn{6}{l}{\textit{HyenaDNA (Long Convolutions + NTP + last-token; Reference-Genome DS)}} \\
\midrule
GUE TF Binding & \textbf{0.694} $\pm$ .006 & \textbf{0.679} $\pm$ .005 & 0.582 $\pm$ .041 & 0.567 $\pm$ .035 & -16.07\% \\
GUE Promoter & 0.675 $\pm$ .014 & 0.710 $\pm$ .015 & \textbf{0.761} $\pm$ .009 & \textbf{0.730} $\pm$ .009 & +12.82\% \\
GUE Splice Site & 0.515 $\pm$ .011 & 0.342 $\pm$ .008 & \textbf{0.548} $\pm$ .011 & \textbf{0.369} $\pm$ .008 & +6.36\% \\
VB Coding Path. & 0.517 $\pm$ .003 & 0.445 $\pm$ .003 & \textbf{0.548} $\pm$ .010 & \textbf{0.472} $\pm$ .014 & +6.02\% \\
VB Exp. Effect & \textbf{0.577} $\pm$ .003 & \textbf{0.561} $\pm$ .002 & 0.556 $\pm$ .007 & 0.535 $\pm$ .010 & -3.63\% \\
VB Common/Rare & \textbf{0.506} $\pm$ .006 & \textbf{0.504} $\pm$ .005 & 0.499 $\pm$ .005 & 0.498 $\pm$ .006 & -1.35\% \\
VB meQTL & 0.511 $\pm$ .008 & 0.460 $\pm$ .004 & \textbf{0.513} $\pm$ .004 & \textbf{0.462} $\pm$ .006 & +0.35\% \\
VB sQTL & 0.550 $\pm$ .001 & 0.556 $\pm$ .001 & \textbf{0.553} $\pm$ .002 & \textbf{0.555} $\pm$ .002 & +0.58\% \\
LRB Causal eQTL & 0.552 $\pm$ .011 & 0.558 $\pm$ .006 & \textbf{0.628} $\pm$ .004 & \textbf{0.619} $\pm$ .006 & +13.64\% \\
\bottomrule
\end{tabular}
\end{table}

the \textit{Best Baseline GFM} column reports the maximum score achieved by any of these models in their original form for a given task, and \textit{Best JEPA-DNA} the peak performance reached by extending three of them with our framework, so the baseline column is an oracle over seven leading models against only three JEPA models, making this a stricter test of our approach. Under this linear probing protocol, JEPA-DNA establishes a new performance ceiling across the majority of tasks, spanning regulatory syntax to long-range causal effects, consistently outperforming the best-performing individual baselines and thereby elevating the representational limits of current genomic architectures.

\paragraph{Zero-Shot Performance for Additional Backbones.}\label{subsec:appendix_zs}
Tables~\ref{tab:zs_continual_baselines} and~\ref{tab:zs_detailed} provide the full zero-shot results for DNABERT-2, 
Nucleotide Transformer v3, and HyenaDNA, extending the analysis presented in the main text. These results confirm that the performance gains observed in zero-shot tasks are consistent across disparate model architectures, pre-training objectives and datasets.

\begin{table}[h!]
\centering
\scriptsize
\renewcommand{\arraystretch}{1.2}
\setlength{\tabcolsep}{2pt}
\caption{AUROC comparison of supervised tasks for genomic foundation models. Multiple backbone models are evaluated under a linear probing setup on diverse genomic prediction benchmarks to identify the current state of the art for each task under these settings.}
\label{tab:baselines}
\begin{tabular}{@{}l c c c c c c c@{}}
\toprule
\textbf{Task} & \textbf{DNABERT} & \textbf{DNABERT-2} & \textbf{NTv3 (8M)} & \textbf{NTv3 (100M)} & \textbf{EVO 2 (1B)} & \textbf{HyenaDNA} & \textbf{Caduceus} \\
\midrule
GUE TF Binding          & 0.812 & 0.788 & 0.792 & \textbf{0.837} & 0.630 & 0.698 & 0.651 \\
GUE Promoter            & 0.836 & 0.884 & 0.904 & \textbf{0.907} & 0.815 & 0.686 & 0.693 \\
GUE Splice Site         & 0.574 & 0.653 & 0.640 & \textbf{0.713} & 0.657 & 0.506 & 0.418 \\
VB Coding Pathogenicity & 0.613 & 0.624 & 0.595 & \textbf{0.668} & 0.659 & 0.515 & 0.586 \\
% VB Non-coding Pathogen. & \textbf{0.846} & 0.605 & 0.594 & 0.625 & 0.659 & 0.537 & 0.553 \\
VB Expression Effect    & 0.666 & \textbf{0.674} & 0.617 & 0.642 & 0.628 & 0.580 & 0.632 \\
VB Common vs. Rare      & \textbf{0.589} & 0.505 & 0.466 & 0.502 & 0.525 & 0.503 & 0.501 \\
VB meQTL                & \textbf{0.635} & 0.623 & 0.541 & 0.576 & 0.560 & 0.503 & 0.453 \\
VB sQTL                 & 0.558 & 0.585 & 0.580 & \textbf{0.591} & 0.573        & 0.550 & 0.572 \\
LRB Causal eQTL         & 0.694 & \textbf{0.697} & 0.672 & 0.690 & 0.662 & 0.565 & 0.665 \\
\bottomrule
\end{tabular}
\end{table}
\begin{comment}
\begin{table}[h!]
\centering
\small
\renewcommand{\arraystretch}{1.2}
\caption{Performance comparison of JEPA-DNA models against current state-of-the-art baselines. For each supervised genomic task, we report the best AUROC achieved by any baseline foundation model and the best AUROC obtained by a JEPA-augmented model, indicating whether JEPA-DNA attains state-of-the-art performance.}
\label{tab:jepa_vs_sota}
\begin{tabular}{@{}l c c c@{}}
\toprule
\textbf{Task} & \textbf{Current SOTA AUROC} & \textbf{Best JEPA AUROC} & \textbf{JEPA is SOTA} \\
\midrule
GUE TF Binding          & 0.837 & 0.849 & V \\
GUE Promoter            & 0.907 & 0.937 & V \\
GUE Splice Site         & 0.713 & 0.733 & V \\
\hdashline
VB Coding Pathogenicity & 0.668 & 0.675 & V \\
VB Non-coding Pathogen. & 0.846 & 0.650 & X \\
VB Expression Effect    & 0.674 & 0.689 & V \\
VB Common vs. Rare      & 0.589 & 0.510 & X \\
VB meQTL                & 0.635 & 0.645 & V \\
VB sQTL                 & 0.591 & 0.590 & X \\
\hdashline
LRB Causal eQTL         & 0.697 & 0.711 & V \\
\bottomrule
\end{tabular}
\end{table}
\end{comment}

\begin{table}[h!]
\centering
%\small
\scriptsize
\setlength{\tabcolsep}{6pt}
\caption{Linear probing AUROC comparison between the top-performing GFMs and their corresponding JEPA-DNA augmented counterparts. Bold values indicate the best result under this protocol.}
\label{tab:jepa_vs_sota}
%\resizebox{0.6\textwidth}{!}{%
\begin{tabular}{@{}l r cc@{}}
\toprule
\textbf{Task} & \textbf{Len.} & \textbf{Best Baseline GFM} & \textbf{Best JEPA-DNA (Ours)} \\
\midrule
TF Binding              & 100  & 0.837 & \textbf{0.849} \\
Promoter                & 300  & 0.907 & \textbf{0.937} \\
Splice Site             & 400  & 0.713 & \textbf{0.733} \\
Coding Pathogenicity    & 1024 & 0.668 & \textbf{0.675} \\
% Non-coding Pathogen.    & 1024 & \textbf{0.846} & 0.650 \\
Expression Effect       & 1024 & 0.674 & \textbf{0.689} \\
Common vs. Rare         & 1024 & \textbf{0.589} & 0.510 \\
meQTL                   & 1024 & 0.635 & \textbf{0.645} \\
sQTL                    & 1024 & \textbf{0.591} & 0.590 \\
Causal eQTL             & 8192 & 0.697 & \textbf{0.711} \\
\bottomrule
\end{tabular}
%}
\end{table}

\begin{table}[tbh!]
\centering
\scriptsize
\renewcommand{\arraystretch}{1.1}
\caption{Zero-shot variant-effect performance (via AUROC) across all backbones, compared against the original checkpoint and the two continually-trained native baselines. \textit{Step-matched} and \textit{Compute-matched} continue training each backbone with its own native objective and dataset for the same number of iterations, and for the same amount of compute time, as JEPA-DNA respectively. Gain denotes the \emph{relative} AUROC gain of JEPA-DNA over the original checkpoint / step-matched / compute-matched models. Bold values indicate the top performance within each backbone group.}
\label{tab:zs_continual_baselines}
{\color{black}
\begin{tabular}{@{}l cccc r@{}}
\toprule
\textbf{Task} & \textbf{Original} & \textbf{Step-} & \textbf{Compute-} & \textbf{JEPA-DNA} & \textbf{Gain} \\
              & \textbf{checkpoint} & \textbf{matched} & \textbf{matched} & \textbf{(Ours)} & \scriptsize \% AUROC \\
\midrule
\multicolumn{6}{l}{\textit{DNABERT-2 (Transformer + MLM + CLS; Multi-Species DS)}} \\
\midrule
BEND Expression Effect   & 0.543 & 0.530 & 0.536 & \textbf{0.574} & +5.71 / +8.21 / +7.00 \\
BEND Disease Variant     & 0.680 & 0.638 & 0.631 & \textbf{0.762} & +12.06 / +19.48 / +20.68 \\
TraitGym Complex         & \textbf{0.517} & 0.512 & 0.514 & 0.503 & -2.71 / -1.80 / -2.06 \\
TraitGym Mendelian       & 0.515 & 0.530 & 0.526 & \textbf{0.540} & +4.85 / +1.93 / +2.57 \\
Songlab ClinVar          & \textbf{0.540} & 0.518 & 0.516 & 0.527 & -2.41 / +1.72 / +2.15 \\
LOL-EVE Causal eQTL      & 0.515 & 0.522 & 0.523 & \textbf{0.545} & +5.83 / +4.47 / +4.26 \\
BRCA1                    & 0.505 & 0.522 & 0.501 & \textbf{0.527} & +4.36 / +0.93 / +5.12 \\
LRB Pathogenic OMIM      & 0.537 & 0.516 & 0.503 & \textbf{0.565} & +5.21 / +9.49 / +12.43 \\
\midrule
\multicolumn{6}{l}{\textit{Nucleotide Transformer v3 (Hybrid + MLM + Mean Pooling; Open-Genome 2 DS)}} \\
\midrule
BEND Expression Effect   & 0.514 & 0.513 & 0.512 & \textbf{0.579} & +12.65 / +12.97 / +12.99 \\
BEND Disease Variant     & 0.713 & 0.716 & \textbf{0.716} & 0.680 & -4.63 / -4.96 / -4.99 \\
TraitGym Complex         & 0.495 & 0.480 & 0.508 & \textbf{0.521} & +5.25 / +8.46 / +2.61 \\
TraitGym Mendelian       & 0.507 & 0.497 & 0.514 & \textbf{0.542} & +6.90 / +8.97 / +5.50 \\
Songlab ClinVar          & 0.473 & 0.479 & 0.472 & \textbf{0.504} & +6.55 / +5.11 / +6.86 \\
LOL-EVE Causal eQTL      & 0.537 & 0.536 & 0.536 & \textbf{0.545} & +1.49 / +1.62 / +1.68 \\
BRCA1                    & 0.628 & 0.625 & \textbf{0.636} & 0.599 & -4.62 / -4.09 / -5.87 \\
LRB Pathogenic OMIM      & 0.505 & 0.511 & 0.497 & \textbf{0.533} & +5.54 / +4.27 / +7.19 \\
\midrule
\multicolumn{6}{l}{\textit{HyenaDNA (Long Convolutions + NTP + last-token; Reference-Genome DS)}} \\
\midrule
BEND Expression Effect   & 0.486 & 0.487 & 0.501 & \textbf{0.535} & +10.08 / +9.91 / +6.87 \\
BEND Disease Variant     & 0.504 & 0.507 & \textbf{0.512} & 0.502 & -0.40 / -0.93 / -1.96 \\
TraitGym Complex         & 0.503 & 0.492 & 0.501 & \textbf{0.508} & +0.99 / +3.34 / +1.41 \\
TraitGym Mendelian       & 0.496 & 0.518 & 0.513 & \textbf{0.548} & +10.48 / +5.87 / +6.79 \\
Songlab ClinVar          & 0.497 & 0.500 & \textbf{0.513} & 0.502 & +1.01 / +0.42 / -2.22 \\
LOL-EVE Causal eQTL      & \textbf{0.559} & 0.548 & 0.522 & 0.539 & -3.58 / -1.65 / +3.23 \\
BRCA1                    & 0.511 & 0.496 & \textbf{0.528} & 0.483 & -5.48 / -2.68 / -8.61 \\
LRB Pathogenic OMIM      & 0.507 & 0.506 & \textbf{0.511} & 0.489 & -3.55 / -3.43 / -4.24 \\
\bottomrule
\end{tabular}
}
\end{table}

\begin{table}[h!]
\centering
\scriptsize
\renewcommand{\arraystretch}{1.1}
\caption{Zero-shot capabilities of JEPA-DNA augmented models against their original baselines. Bold values denote superior zero-shot performance within each backbone group. Gain denotes the \emph{relative} AUROC gain over the corresponding original checkpoint baseline.}
\label{tab:zs_detailed}
\begin{tabular}{@{}l cc cc r@{}}
\toprule
\textbf{Task} & \multicolumn{2}{c}{\textbf{Baseline}} & \multicolumn{2}{c}{\textbf{JEPA-DNA (Ours)}} & \textbf{Gain} \\
\cmidrule(lr){2-3} \cmidrule(lr){4-5} \cmidrule(lr){6-6}
& \scriptsize AUROC & \scriptsize AUPRC & \scriptsize AUROC & \scriptsize AUPRC & \scriptsize \% AUROC \\
\midrule
\multicolumn{6}{l}{\textit{DNABERT-2 (Transformer + MLM + CLS; Multi-Species DS)}} \\
\midrule
BEND Expression Effect & 0.543 & 0.084 & \textbf{0.574} & \textbf{0.091} & +5.71\% \\
BEND Disease Variant   & 0.680 & 0.141 & \textbf{0.762} & \textbf{0.262} & +12.06\% \\
TraitGym Complex       & \textbf{0.517} & \textbf{0.106} & 0.503 & 0.100 & -2.71\% \\
TraitGym Mendelian     & 0.515 & 0.103 & \textbf{0.540} & \textbf{0.110} & +4.85\% \\
% VEP-eval ClinVar       & \textbf{0.563} & \textbf{0.347} & 0.551 & 0.338 & -2.13\% \\
Songlab ClinVar        & \textbf{0.540} & \textbf{0.566} & 0.527 & 0.557 & -2.41\% \\
% Indels ClinVar         & 0.571 & 0.880 & \textbf{0.588} & \textbf{0.887} & +2.98\% \\
LOL-EVE Causal eQTL    & 0.515 & 0.083 & \textbf{0.545} & \textbf{0.090} & +5.83\% \\
BRCA1                  & 0.505 & 0.210 & \textbf{0.527} & \textbf{0.218} & +4.36\% \\
LRB Pathogenic OMIM    & 0.537 & 0.002 & \textbf{0.565} & 0.002 & +5.21\% \\
\midrule
\multicolumn{6}{l}{\textit{Nucleotide Transformer v3 (Hybrid + MLM + Mean Pooling; Open-Genome 2 DS)}} \\
\midrule
BEND Expression Effect & 0.514 & 0.079 & \textbf{0.579} & \textbf{0.091} & +12.65\% \\
BEND Disease Variant   & \textbf{0.713} & \textbf{0.147} & 0.680 & 0.131 & -4.63\% \\
TraitGym Complex       & 0.495 & 0.099 & \textbf{0.521} & \textbf{0.104} & +5.25\% \\
TraitGym Mendelian     & 0.507 & 0.106 & \textbf{0.542} & \textbf{0.110} & +6.90\% \\
% VEP-eval ClinVar       & \textbf{0.682} & \textbf{0.522} & 0.681 & 0.492 & -0.15\% \\
Songlab ClinVar        & 0.473 & 0.523 & \textbf{0.504} & \textbf{0.558} & +6.55\% \\
% Indels ClinVar         & \textbf{0.654} & \textbf{0.904} & 0.569 & 0.880 & -13.00\% \\
LOL-EVE Causal eQTL    & 0.537 & 0.084 & \textbf{0.545} & \textbf{0.087} & +1.49\% \\
BRCA1                  & \textbf{0.628} & \textbf{0.359} & 0.599 & 0.290 & -4.62\% \\
LRB Pathogenic OMIM    & 0.505 & 0.002 & \textbf{0.533} & 0.002 & +5.54\% \\
\midrule
\multicolumn{6}{l}{\textit{Long Convolutions + NTP + last-token; Reference-Genome DS}} \\
\midrule
BEND Expression Effect & 0.486 & 0.072 & \textbf{0.535} & \textbf{0.084} & +10.08\% \\
BEND Disease Variant   & \textbf{0.504} & \textbf{0.075} & 0.502 & 0.074 & -0.40\% \\
TraitGym Complex       & 0.503 & 0.099 & \textbf{0.508} & \textbf{0.102} & +0.99\% \\
TraitGym Mendelian     & 0.496 & 0.103 & \textbf{0.548} & \textbf{0.124} & +10.48\% \\
% VEP-eval ClinVar       & 0.501 & 0.307 & \textbf{0.509} & \textbf{0.311} & +1.60\% \\
Songlab ClinVar        & 0.497 & 0.542 & \textbf{0.502} & \textbf{0.543} & +1.01\% \\
% Indels ClinVar         & 0.520 & 0.865 & \textbf{0.524} & \textbf{0.866} & +0.77\% \\
LOL-EVE Causal eQTL    & \textbf{0.559} & \textbf{0.089} & 0.539 & 0.086 & -3.58\% \\
BRCA1                  & \textbf{0.511} & \textbf{0.216} & 0.483 & 0.200 & -5.48\% \\
LRB Pathogenic OMIM    & \textbf{0.507} & 0.002 & 0.489 & 0.002 & -3.55\% \\
\bottomrule
\end{tabular}
\end{table}

\begin{table}[h!]
\centering
\scriptsize
\setlength{\tabcolsep}{3pt}
\renewcommand{\arraystretch}{1.1}
\caption{Comparative analysis of training paradigms. All JEPA variants include the VICReg loss. When the token type is JEPA, the JEPA objective is applied to the representations of individual masked tokens, analogous to the I-JEPA formulation. Bold values indicate the optimal configuration. Gains are recomputed against the revised continual-training runs and are consistent with Table~\ref{tab:ablation_jepa_dna}. Gain denotes the \emph{relative} AUROC gain over the corresponding baseline.}
\label{tab:ablation_training_obj}
\begin{tabular}{@{}llccccccccr@{}}
\toprule
\textbf{Paradigm} & \textbf{Token} & \textbf{Global} & \textbf{Sched.} & \multicolumn{2}{c}{\textbf{Promoter}} & \multicolumn{2}{c}{\textbf{Splice Site}} & \multicolumn{2}{c}{\textbf{Expression}} & \textbf{Avg. Gain} \\
\cmidrule(lr){5-6} \cmidrule(lr){7-8} \cmidrule(lr){9-10}
& \textbf{Type} & \textbf{JEPA} & \textbf{Mask} & \scriptsize AUROC & \scriptsize AUPRC & \scriptsize AUROC & \scriptsize AUPRC & \scriptsize AUROC & \scriptsize AUPRC & \scriptsize{\% AUROC} \\
\midrule
\textit{MLM Only} \\
MLM  & MLM  & $\times$ & $\times$      & 0.929 & 0.914 & 0.674 & 0.467 & 0.532 & 0.084 & +2.09\% \\
MLM  & MLM  & $\times$ & $\checkmark$  & 0.927 & 0.911 & 0.669 & 0.463 & 0.530 & 0.084 & +1.67\% \\
\midrule
\textit{JEPA Only} \\
JEPA & JEPA & $\checkmark$ & $\checkmark$      & 0.919 & 0.906 & 0.680 & 0.474 & 0.562 & 0.089 & +3.86\% \\
\midrule
\textit{Hybrid (JEPA-DNA, \textbf{Ours})} \\
JEPA & MLM  & $\checkmark$ & $\times$      & 0.928 & 0.908 & 0.689 & 0.482 & 0.565 & 0.088 & +4.85\% \\
\textbf{JEPA} & \textbf{MLM} & \textbf{$\checkmark$} & $\checkmark$ & \textbf{0.928} & \textbf{0.913} & \textbf{0.705} & \textbf{0.500} & \textbf{0.574} & \textbf{0.091} & \textbf{+6.22\%} \\
\bottomrule
\end{tabular}
\end{table}
\begin{comment}
\begin{table}[h!]
\centering
\small
\renewcommand{\arraystretch}{1.2}
\caption{Performance comparison of predictor architectures for DNABERT-2 models. Several attention-based predictor architectures with different depths were trained under JEPA continual learning and evaluated on three genomic tasks. The baseline corresponds to the pre-trained DNABERT-2 model without additional JEPA continual training, and the MLP row reports results for a non-attention-based predictor.}
\label{tab:predictors}
\begin{tabular}{@{}l ccc ccc cc c@{}}
\toprule
& \multicolumn{3}{c}{GUE Promoter} & \multicolumn{3}{c}{GUE Splice Site} & \multicolumn{2}{c}{Bend expression} & Overall \\
\cmidrule(lr){2-4} \cmidrule(lr){5-7} \cmidrule(lr){8-9} \cmidrule(lr){10-10}
Predictor & \scriptsize MCC & \scriptsize AUROC & \scriptsize AUPRC & \scriptsize MCC & \scriptsize AUROC & \scriptsize AUPRC & \scriptsize AUROC & \scriptsize AUPRC & \scriptsize Mean Gain \\
\midrule
Baseline       & 0.607 & 0.884 & 0.877 & 0.000 & 0.653 & 0.452 & 0.543 & 0.084 & 0.00\% \\
2 layers       & 0.719 & 0.928 & 0.911 & 0.103 & 0.691 & 0.485 & 0.561 & 0.087 & +4.70\% \\
4 layers (ours)& 0.717 & 0.928 & 0.913 & 0.139 & 0.705 & 0.500 & 0.574 & 0.091 & +6.22\% \\
6 layers       & 0.727 & 0.928 & 0.910 & 0.006 & 0.655 & 0.451 & 0.555 & 0.086 & +2.50\% \\
8 layers       & 0.701 & 0.917 & 0.900 & 0.034 & 0.662 & 0.459 & 0.549 & 0.086 & +2.07\% \\
MLP            & 0.646 & 0.900 & 0.889 & 0.012 & 0.648 & 0.450 & 0.564 & 0.088 & +1.64\% \\
\bottomrule
\end{tabular}
\end{table}
\end{comment}

\begin{table}[tbh!]
\centering
\scriptsize
\renewcommand{\arraystretch}{1.1}
\caption{Evaluation of predictor head's architecture and depth. Bold values denote the best-performing configuration. Gain denotes the \emph{relative} AUROC gain over the corresponding baseline, not an absolute AUROC difference.}
\label{tab:predictor_arch}
%\resizebox{0.7\textwidth}{!}{
\begin{tabular}{@{}l cc cc cc r@{}}
\toprule
\textbf{Predictor} & \multicolumn{2}{c}{\textbf{GUE Promoter}} & \multicolumn{2}{c}{\textbf{GUE Splice Site}} & \multicolumn{2}{c}{\textbf{BEND Expression}} & \textbf{Avg. Gain} \\
\cmidrule(lr){2-3} \cmidrule(lr){4-5} \cmidrule(lr){6-7}
\textbf{Architecture} & \scriptsize AUROC & \scriptsize AUPRC & \scriptsize AUROC & \scriptsize AUPRC & \scriptsize AUROC & \scriptsize AUPRC & \scriptsize{\% AUROC } \\
\midrule
Baseline        & 0.884 & 0.877 & 0.653 & 0.452 & 0.543 & 0.084 & --- \\
\midrule
MLP             & 0.900 & 0.889 & 0.648 & 0.450 & 0.564 & 0.088 & +1.64\% \\
\midrule
2 layers        & 0.928 & 0.911 & 0.691 & 0.485 & 0.561 & 0.087 & +4.70\% \\
\textbf{4 layers (Ours)} & \textbf{0.928} & \textbf{0.913} & \textbf{0.705} & \textbf{0.500} & \textbf{0.574} & \textbf{0.091} & \textbf{+6.22\%} \\
6 layers        & 0.928 & 0.910 & 0.655 & 0.451 & 0.555 & 0.086 & +2.50\% \\
8 layers        & 0.917 & 0.900 & 0.662 & 0.459 & 0.549 & 0.086 & +2.07\% \\
\bottomrule
\end{tabular}
%}
\end{table}
\begin{comment}
\begin{table}[tbh!]
\centering
\small
\renewcommand{\arraystretch}{1.2}
\caption{Performance comparison of optimizers for DNABERT-2 models. Several DNABERT-2 models were trained under JEPA-DNA continual learning using different optimization algorithms and evaluated on three genomic tasks. The baseline corresponds to the pre-trained DNABERT-2 model without additional JEPA-DNA continual training.}
\label{tab:optimizers}
\begin{tabular}{@{}l ccc cccc cc c@{}}
\toprule
& \multicolumn{3}{c}{GUE Promoter} & \multicolumn{3}{c}{GUE Splice Site} & \multicolumn{2}{c}{Bend expression} & Overall \\
\cmidrule(lr){2-4} \cmidrule(lr){5-7} \cmidrule(lr){8-9} \cmidrule(lr){10-10}
Optimizer & \scriptsize MCC & \scriptsize AUROC & \scriptsize AUPRC & \scriptsize MCC & \scriptsize AUROC & \scriptsize AUPRC & \scriptsize AUROC & \scriptsize AUPRC & \scriptsize Mean Gain \\
\midrule
Baseline      & 0.607 & 0.884 & 0.877 & 0.000 & 0.653 & 0.452 & 0.543 & 0.084 & 0.00\%   \\
Adam          & 0.391 & 0.776 & 0.777 & 0.035 & 0.523 & 0.351 & 0.518 & 0.080 & -12.24\% \\
AdamW         & 0.407 & 0.799 & 0.795 & -0.012 & 0.589 & 0.388 & 0.576 & 0.091 & -4.45\%  \\
AdaBelief     & 0.626 & 0.901 & 0.894 & 0.014 & 0.662 & 0.454 & 0.545 & 0.086 & 1.22\%   \\
SGDM (ours)   & 0.717 & 0.928 & 0.913 & 0.139 & 0.705 & 0.500 & 0.574 & 0.091 & 6.22\%   \\
\bottomrule
\end{tabular}
\end{table}
\end{comment}
\begin{table}[tbh!]
\centering
\scriptsize
\renewcommand{\arraystretch}{1.1}
\caption{Optimizer Sensitivity in JEPA-DNA Continual Pre-training. We evaluate the impact of different optimization algorithms on the stability and performance of JEPA-DNA using DNABERT-2 as the backbone. Bold values indicate peak performance. Gain denotes the \emph{relative} AUROC gain over the corresponding baseline, not an absolute AUROC difference.}
\label{tab:optimizers}
\begin{tabular}{@{}l cc cc cc r@{}}
\toprule
\textbf{Optimizer} & \multicolumn{2}{c}{\textbf{Promoter}} & \multicolumn{2}{c}{\textbf{Splice Site}} & \multicolumn{2}{c}{\textbf{Expression}} & \textbf{Avg. Gain} \\
\cmidrule(lr){2-3} \cmidrule(lr){4-5} \cmidrule(lr){6-7}
& \scriptsize AUROC & \scriptsize AUPRC & \scriptsize AUROC & \scriptsize AUPRC & \scriptsize AUROC & \scriptsize AUPRC & \scriptsize{\% AUROC} \\
\midrule
Baseline        & 0.884 & 0.877 & 0.653 & 0.452 & 0.543 & 0.084 & --- \\
\midrule
Adam~\cite{kingma2014adam}            & 0.776 & 0.777 & 0.523 & 0.351 & 0.518 & 0.080 & -12.24\% \\
AdamW~\cite{loshchilov2017decoupled}           & 0.799 & 0.795 & 0.589 & 0.388 & 0.576 & 0.091 & -4.45\%  \\
AdaBelief~\cite{zhuang2020adabelief}       & 0.901 & 0.894 & 0.662 & 0.454 & 0.545 & 0.086 & +1.22\%  \\
\textbf{SGDM~\cite{bottou2010large}} & \textbf{0.928} & \textbf{0.913} & \textbf{0.705} & \textbf{0.500} & \textbf{0.574} & \textbf{0.091} & \textbf{+6.22\%} \\
\bottomrule
\end{tabular}
\end{table}

\paragraph{Extended Ablation Studies.}\label{subsec:appendix_ablation}
Table~\ref{tab:ablation_jepa_dna} in the main text isolates the contribution of each component of the framework. For the compute-matched budget, native continual MLM, with neither an EMA target nor a latent objective, yields only $+2.16\%$, so roughly $65\%$ of the full $+6.22\%$ gain originates from components beyond matched continual training. Removing the latent objective while retaining regularization and scheduled masking (MLM+VR+SM) reaches $+3.24\%$, so regularization clearly contributes, but adding latent prediction approximately doubles the gain. Conversely, the latent objective without any token-level loss (JEPA+VR+SM) already attains $+3.85\%$, exceeding every MLM-only variant, which indicates that the two objectives are complementary rather than redundant. Neither the EMA target nor scheduled masking accounts for the gain in isolation, as EMA leaves matched continual MLM essentially unchanged ($+2.16\%$ to $+2.09\%$) while scheduled masking pays only when coupled with the latent objective ($+4.85\%$ to $+6.22\%$). Finally, MLM+JEPA without VICReg falls below MLM alone ($+1.63\%$ against $+2.16\%$). Absent a variance term the predictor drives the context encoder toward a low-rank solution, so VICReg is not an independent booster but the enabling condition for latent prediction.

Table~\ref{tab:ablation_training_obj} compares training paradigms according to the granularity at which each objective is applied. All JEPA variants share the same global objective on the sequence representative and differ only in the loss applied at the token level. Extending the latent objective to the token level as well, so that the individual masked tokens are also predicted in latent space and no MLM loss is used, reaches $+3.86\%$, whereas retaining the native MLM loss at the token level attains $+6.22\%$. The native token-level objective therefore remains important: where the tokenization provides a discrete vocabulary, classifying the masked tokens is a stronger local signal than reconstructing their representations in latent space. The two objectives are best combined by keeping MLM at the token level and applying JEPA to the sequence representative, rather than by using latent prediction at both granularities.

Table~\ref{tab:predictor_arch} evaluates the predictor head. An attention-based predictor outperforms the MLP variant, and performance peaks at a depth of four layers.

All previous ablation tables (Tables~\ref{tab:ablation_jepa_dna},~\ref{tab:ablation_training_obj} and~\ref{tab:predictor_arch}) use the SGDM optimizer, so optimizer choice is held fixed across them. In Table~\ref{tab:optimizers} we compare optimization algorithms and find that adaptive methods are harmful in this continual-training regime, with Adam and AdamW degrading performance relative to the baseline ($-12.24\%$ and $-4.45\%$) and AdaBelief yielding only a marginal gain ($+1.22\%$), whereas SGDM realizes the full $+6.22\%$.

\paragraph{DNASimCLR Comparison.} \label{dnasimcler_app}
We compare our JEPA-DNA approach against DNASimCLR \citep{yang2024dnasimclr}, a contrastive learning framework based on SimCLR \citep{chen2020simple}. Due to the absence of publicly available code, we reimplement the method following the original description and refine the official SimCLR implementation accordingly, adapting it to the DNA domain. Specifically, DNA sequences are encoded as one-hot representations and processed using a ResNet50 backbone, treating sequences as image-like inputs. Training uses augmented pairs generated by randomly masking approximately 30\% of each sequence, where positive pairs originate from the same sequence and negatives are drawn from other sequences within the batch. To ensure a fair comparison of training paradigms, we adopt a multi-species dataset and sequence lengths consistent with DNABERT-2. As shown in Table~\ref{tab:dnasimclr}, JEPA-DNA with a DNABERT-2 backbone consistently outperforms DNASimCLR across all evaluated tasks, with particularly notable gains on splice site prediction, resulting in an average AUROC improvement of 7.57\%. Because the published configuration differs from ours in backbone, initialization and training history, we further report three controlled variants. Increasing the contrastive vision backbone from ResNet-50 with 25.6M parameters to ResNet-101-2 with 126.9M parameters, beyond the 117M parameters of the DNABERT-2 backbone used by JEPA-DNA, does not increase performance ($-1.53\%$). Replacing it with the same pretrained DNABERT-2 backbone used by JEPA-DNA, while retaining the contrastive loss and adapting the vision-style augmentations to token-level operations, recovers $+4.06\%$ but remains below JEPA-DNA ($+7.57\%$). Finally, adding the MLM objective to the contrastive one degrades it to $-0.34\%$, whereas the same addition to the latent objective improves it (Table~\ref{tab:ablation_jepa_dna}).

\begin{table}[tbh!]
\centering
\scriptsize
\renewcommand{\arraystretch}{1.1}
\caption{DNASimCLR vs. JEPA-DNA across genomic tasks. We additionally report three controlled variants: scaling the contrastive vision backbone (ResNet-101-2), replacing it with the same pretrained DNABERT-2 backbone used by JEPA-DNA, and adding the MLM objective to the contrastive one. Gain denotes the \emph{relative} AUROC gain over the published DNASimCLR configuration.}
\label{tab:dnasimclr}
\begin{tabular}{@{}l cc cc cc r@{}}
\toprule
\textbf{Learning Paradigm} & \multicolumn{2}{c}{\textbf{Promoter}} & \multicolumn{2}{c}{\textbf{Splice Site}} & \multicolumn{2}{c}{\textbf{Expression}} & \textbf{Avg. Gain} \\
\cmidrule(lr){2-3} \cmidrule(lr){4-5} \cmidrule(lr){6-7}
& \scriptsize AUROC & \scriptsize AUPRC & \scriptsize AUROC & \scriptsize AUPRC & \scriptsize AUROC & \scriptsize AUPRC & \scriptsize{\% AUROC}  \\
\midrule
DNASimCLR~\cite{yang2024dnasimclr}        & 0.922 & 0.912 & 0.600 & 0.407 & 0.549 & 0.085 & --- \\
DNASimCLR (ResNet-101-2, 126.9M) & 0.917 & 0.909 & 0.579 & 0.391 & 0.546 & 0.084 & -1.53\% \\
DNASimCLR w/ pretrained DNABERT-2 (117M) & 0.921 & 0.919 & 0.680 & 0.500 & 0.543 & 0.071 & +4.06\% \\
DNASimCLR w/ DNABERT-2 + MLM & 0.921 & 0.911 & 0.686 & 0.483 & 0.465 & 0.070 & -0.34\% \\
% \midrule
\textbf{JEPA-DNA (Ours)} & \textbf{0.928} & \textbf{0.913} & \textbf{0.705} & \textbf{0.500} & \textbf{0.574} & \textbf{0.091} & \textbf{+7.57\%} \\
\bottomrule
\end{tabular}
\end{table}

\paragraph{Embeddings Analysis.} \label{subsec:appendix_umap}
We defined a set of genomic loci on human chromosome 22 (GRCh38) using public Ensembl annotations \citep{harrison2024ensembl}, including the gene annotation (GTF), regulatory feature track (GFF3), and motif feature track (GFF3) from the same release. From the GTF, we retained protein-coding and long non-coding RNA genes, representing each gene by the midpoint of its annotated span. Splice donor and acceptor sites were inferred from consecutive exons within the same transcript via intron boundaries. Regulatory and motif intervals were extracted from GFF3 files, summarized by their midpoints, and labeled by feature type (e.g., promoter, enhancer, CTCF binding, open chromatin, transcription factor binding). This process yielded approximately 37,000 labeled windows across nine functional classes on chromosome 22. DNA sequence windows centered on these loci were extracted from the GRCh38 reference and passed through both the JEPA target encoder and context encoder for downstream embedding analysis. For visualization and quantitative evaluation, embeddings were projected to two dimensions using UMAP (Fig. \ref{fig:umap}), with a subsequent balanced per-class subsampling at 85\% of the minimum class size (n = 702, nine biotypes, seed 42). A k-nearest-neighbor classifier applied in the original embedding space with stratified 5-fold cross-validation achieved a weighted F1 score of 0.903, while the context-encoder under the same protocol reached 0.877, indicating slightly weaker class separability compared to the target-encoder.

\begin{figure}
    \centering
    \includegraphics[scale=0.5]{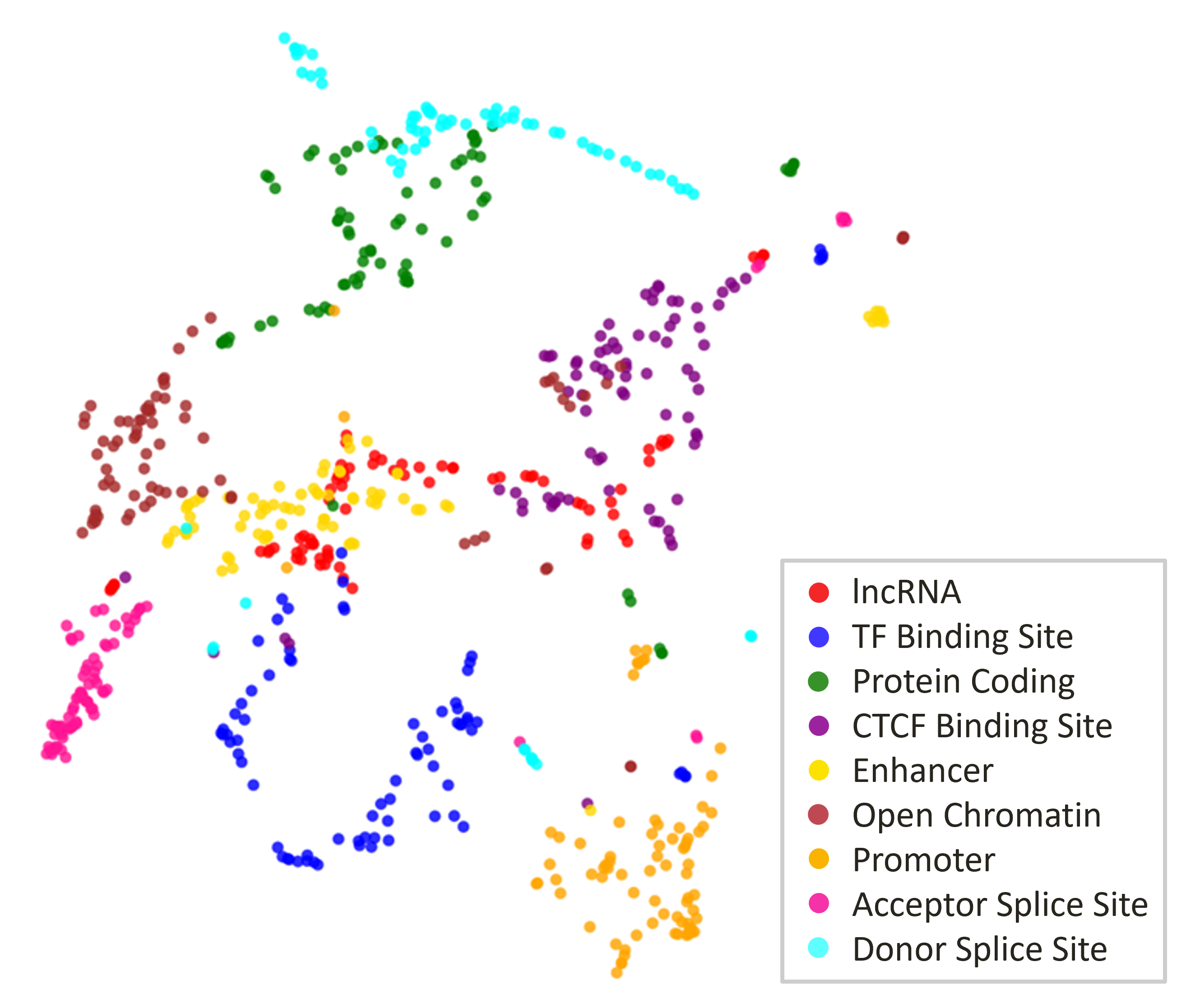}
    \caption{JEPA-DNA Embedding visualization colored by Ensembl annotations.}
    \label{fig:umap}
\end{figure}

\paragraph{Full Fine-Tuning.}\label{subsec:appendix_finetune}
All results in this work follow the frozen linear-probing and zero-shot protocols, which isolate the quality of the learned representation. For completeness, Table~\ref{tab:finetune} reports an end-to-end fine-tuning comparison on DNABERT-2. JEPA-DNA remains ahead on all four tasks, though on the saturated GUE benchmarks (0.90--0.98) the margin narrows to under $0.005$ absolute AUROC, whereas on the non-saturated LRB causal eQTL task it reaches $+8.5\%$ relative.

\begin{table}[tbh!]
\centering
\scriptsize
\renewcommand{\arraystretch}{1.1}
\caption{Full fine-tuning of DNABERT-2 with and without JEPA-DNA continual training (AUROC), reporting both the absolute difference and the relative gain. The GUE tasks are saturated (0.90--0.98), where end-to-end fine-tuning largely washes out differences between the frozen encoders; on the non-saturated LRB causal eQTL task the advantage of the JEPA-DNA representation is retained.}
\label{tab:finetune}
{\color{black}
\begin{tabular}{@{}l cc cr@{}}
\toprule
\textbf{Task} & \textbf{Baseline} & \textbf{JEPA-DNA}  & \textbf{Rel. Gain} \\
\midrule
GUE TF Binding & 0.9025 & \textbf{0.9073} & +0.53\% \\
GUE Promoter (all) & 0.9818 & \textbf{0.9843} & +0.25\% \\
GUE Splice Site & 0.9738 & \textbf{0.9744} & +0.07\% \\
LRB Causal eQTL & 0.6669 & \textbf{0.7237} & +8.51\% \\
\bottomrule
\end{tabular}
}
\end{table}

\paragraph{Behaviour Beyond the Training Context.}\label{subsec:appendix_length}
A known limitation of GFMs is degradation when evaluated on inputs longer than the context they were pre-trained on. In this sequence-length stability analysis, DNABERT-2 (self-supervised context of 512 tokens) and its JEPA-DNA counterpart are evaluated at increasing input lengths (beyond 2,500 bp) on BRCA1 and Enhancer--Target~\citep{cheng2025dnalongbench} (Table~\ref{tab:length_sweep}). On Enhancer--Target, JEPA-DNA leads at every length ($+4.8\%$ to $+7.9\%$ relative). On BRCA1 the two are on par at 4{,}096\,bp, after which the baseline loses 0.215 AUROC by 32{,}768\,bp against 0.058 for JEPA-DNA, a $3.7\times$ slower degradation.

\begin{table}[tbh!]
\centering
\scriptsize
\renewcommand{\arraystretch}{1.1}
\caption{Sequence-length sweep for DNABERT-2 (self-supervised training context of 512 tokens) evaluated at input lengths well beyond its pre-training window (AUROC). We present this as a stability and failure-mode analysis rather than as evidence of long-range reasoning.}
\label{tab:length_sweep}
{\color{black}
\begin{tabular}{@{}ll cccc@{}}
\toprule
\textbf{Task} & \textbf{Model} & \textbf{4{,}096 bp} & \textbf{8{,}192 bp} & \textbf{16{,}384 bp} & \textbf{32{,}768 bp} \\
\midrule
BRCA1 & Baseline & 0.5485 & 0.4468 & 0.4864 & 0.3333 \\
 & JEPA-DNA & 0.5470 & \textbf{0.5280} & \textbf{0.4881} & \textbf{0.4892} \\
\midrule
Enhancer--Target & Baseline & 0.4467 & 0.4817 & 0.6077 & 0.6071 \\
 & JEPA-DNA & \textbf{0.4683} & \textbf{0.5179} & \textbf{0.6546} & \textbf{0.6548} \\
\bottomrule
\end{tabular}
}
\end{table}

\paragraph{Loss-Weight Sensitivity.}\label{subsec:appendix_loss_weights}
Table~\ref{tab:loss_weights} reports the sensitivity of JEPA-DNA to the loss coefficients, perturbing each by $\pm 50\%$ around its default while holding the others fixed, on GUE Splice Site. The tuned configuration remains the strongest, but the method is not brittle: every perturbed setting still exceeds both the original backbone (0.653) and the continual MLM-only baseline (0.670), and the spread across all eight perturbations is 0.021 AUROC.

\subsection{Extended Implementation Details}\label{subsec:appendix_impl_details}
\paragraph{GFM Backbones Details.} We demonstrate the JEPA-DNA framework across multiple genomic backbone models:
\begin{itemize}
\item \textbf{DNABERT-2~\citep{zhou2024dnabert}:} A bidirectional encoder based on the BERT architecture (12 layers, 768 hidden dim, 12 heads, $\sim$117M parameters). It utilizes Byte Pair Encoding (BPE) and supports sequences up to 512 tokens. Sequence representations are derived from the \texttt{[CLS]} token.
\item \textbf{NTv3~\citep{boshar2025foundational}:} A $\sim$100M-parameter hybrid model featuring a convolutional U-Net tower and a 6-layer Transformer stack. It is trained via MLM and utilizes a single nucleotide tokenization. Sequence-level representations are obtained via mean pooling over all token embeddings.
\item \textbf{HyenaDNA~\citep{nguyen2023hyenadna}:} An autoregressive model ($\sim$600K parameters) utilizing long-convolutional blocks (Hyena operators) instead of attention. It is pre-trained via Next-Token Prediction (NTP) and derives sequence representations from the last \texttt{[SEP]} token embedding.\end{itemize}

\paragraph{JEPA-DNA Predictor Configuration.} The JEPA predictor is designed to be lightweight relative to the backbone to prevent a capacity bypass. For Transformer-based backbones, a lightweight 4-layer Transformer encoder operating in a reduced latent space whose dimensionality is set to one half of the backbone hidden size, with 3 attention heads. The predictor employs a pre-norm architecture with GELU activations and frozen sinusoidal positional embeddings. Input embeddings from the context encoder are first projected to the predictor dimension, with masked target positions replaced by learnable mask tokens before adding positional embeddings following the re-masking strategy described in Section~\ref{subsec:masking}. The predictor output for the sequence representative is then projected back to the original backbone hidden dimension for computing the prediction loss against the target encoder representation. For backbones where the sequence representation is obtained via mean pooling, we instead apply a 3-layer MLP predictor directly to the pooled sequence embedding, so that the sequence representative itself is processed by the predictor rather than being implicitly aggregated through attention.

Fig.~\ref{fig:arch_dnabert2} illustrates an instantiation of the JEPA-DNA architecture when using a CLS-based representation. 

\begin{figure}
    \centering
    \includegraphics[scale=0.28]{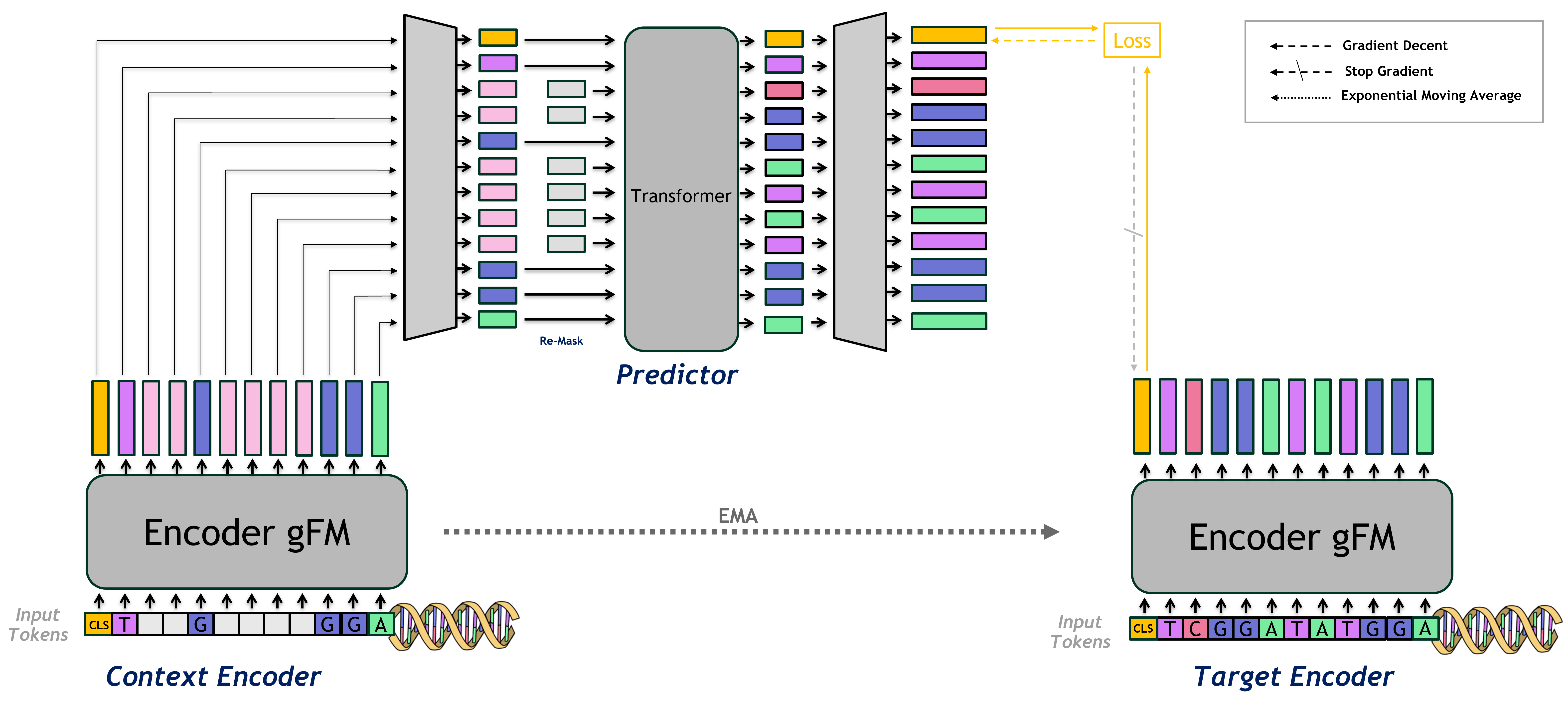}
    \caption{The JEPA-DNA architecture using a CLS-based representation, as implemented with DNABERT-2.}
    \label{fig:arch_dnabert2}
\end{figure}

\begin{table}[tbh!]
\centering
\scriptsize
\setlength{\tabcolsep}{4pt}
\renewcommand{\arraystretch}{1.1}
\caption{Sensitivity to the loss weights on GUE Splice Site (AUROC). Each coefficient is perturbed by $\pm 50\%$ around its default, one at a time. \emph{Original} denotes the original DNABERT-2 checkpoint, \emph{Step} and \emph{Compute} denote the step-matched and compute-matched continual MLM baselines, and \emph{JEPA-DNA} the reported model with no perturbation of the loss weights.}
\label{tab:loss_weights}
{\color{black}
\resizebox{\textwidth}{!}{%
\begin{tabular}{@{}l ccc cccccccc c@{}}
\toprule
& \multicolumn{3}{c}{Baselines} & \multicolumn{8}{c}{Loss-weight perturbations} & \\
\cmidrule(lr){2-4} \cmidrule(lr){5-12}
& Original & Step & Compute & $\lambda_{\mathrm{cov}}\!\times\!0.5$ & $\lambda_{\mathrm{cov}}\!\times\!1.5$ & $\lambda_{\mathrm{var}}\!\times\!0.5$ & $\lambda_{\mathrm{var}}\!\times\!1.5$ & $\lambda_{\mathrm{JEPA}}\!\times\!0.5$ & $\lambda_{\mathrm{JEPA}}\!\times\!1.5$ & $\lambda_{\mathrm{LLM}}\!\times\!0.5$ & $\lambda_{\mathrm{LLM}}\!\times\!1.5$ & JEPA-DNA \\
\midrule
AUROC & 0.653 & 0.670 & 0.672 & 0.681 & 0.679 & 0.685 & 0.675 & 0.679 & 0.680 & 0.677 & 0.696 & 0.705 \\
\bottomrule
\end{tabular}}
}
\end{table}

\paragraph{Continual Training Datasets.} The JEPA-DNA framework is evaluated under a strict "data-parity" regime, where the continual training phase utilizes the exact same genomic sources as the original backbones' pre-training. This ensures that the framework is evaluated on its ability to extract more structured features from existing data rather than benefiting from additional genomic information.
We construct three continual-training datasets for JEPA-DNA, each aligned with the pre-training corpus of its corresponding backbone. For DNABERT-2, we follow the original multi-species setting and use training samples built from the human reference genome (GRCh38/hg38)~\citep{schneider2017evaluation} together with genomic data from five model organisms: mouse~\citep{church2011modernizing}, zebrafish~\citep{howe2013zebrafish}, fruit fly~\citep{hoskins2015release}, nematode~\citep{c1998genome}, and thale cress~\citep{arabidopsis2000analysis}. All genome sequences are obtained from UCSC genome Browser ~\citep{kent2002human}, restricted to valid nucleotides, and chunked into fixed windows of 3200 bp with 50\% overlap, yielding approximately \(4.76 \times 10^6\) sequences. For NTv3, we use the Open Genome 2 corpus \citep{brixi2026genome}, consisting of approximately \(384 \times 10^6\) training records. Sequences are filtered to valid nucleotides, and each training sample is formed by taking a random crop of up to 8{,}192 nucleotides from a record. For HyenaDNA, we derive approximately \(6 \times 10^5\) training windows from the full hg38 human reference FASTA ~\citep{schneider2017evaluation}, each of length 8{,}192 bp with 50\% overlap between consecutive windows, again restricting to valid nucleotides.

\paragraph{Evaluation Benchmarks.} For downstream evaluation, we test on a diverse suite of supervised and zero-shot tasks derived from established genomic benchmarks \citep{larey2026gfmbench}. Our supervised evaluation includes three standard classification tasks from the GUE benchmark~\citep{zhou2024dnabert}: promoter prediction, transcription factor binding site prediction, and splice site prediction. Additionally, we utilize variant effect prediction tasks from VariantBenchmarks~\citep{medvedev2025biotoken}, covering coding pathogenicity classification, common vs. rare variant identification, and quantitative trait loci (QTL) prediction for expression, methylation (meQTL), and splicing (sQTL) effects. We also evaluate on the causal eQTL task from the Long Range Benchmark (LRB)~\citep{kao2024advancing}, which involves high-context inputs.
For zero-shot evaluation, we assess performance on variant effect prediction without task-specific fine-tuning. This includes the BEND benchmark~\citep{marin2024bend} for expression and disease-associated variants, and TraitGym~\citep{benegas2025benchmarking} for predicting Mendelian and complex traits. Furthermore, we evaluate clinical pathogenicity prediction using the Song-Lab ClinVar dataset~\citep{benegas2025dna} and the non-coding pathogenic OMIM task from LRB~\citep{kao2024advancing}. In addition, we include variant-effect prediction on BRCA1 saturation mutagenesis \citep{findlay2018accurate}, and a LOL-EVE causal eQTL task \citep{shearer2024genomic}, providing further coverage of clinically and functionally relevant variant effect prediction tasks.

\paragraph{Evaluation Protocol.}
We employ two primary evaluation strategies to assess the quality of learned representations:
\begin{itemize}
\item Linear Probing: To isolate the quality of the pre-trained features, we keep the backbone frozen and train only a linear classifier on the extracted sequence representations. Performance is measured using three complementary metrics: (1) Area Under the Receiver Operating characteristic curve (AUROC), which measures the model's ability to discriminate between classes across all classification thresholds; (2) Area Under the Precision-Recall Curve (AUPRC), which is particularly informative for imbalanced datasets common in genomics tasks; and (3) Matthews Correlation Coefficient (MCC), a balanced metric that accounts for all four confusion matrix categories and remains robust even when class distributions are highly skewed.
\item Zero-Shot Inference: We evaluate the model's out-of-the-box semantic understanding by computing the cosine similarity between embeddings of reference and variant representative sequences. For variant effect prediction tasks (e.g., ClinVar or TraitGym tasks), we extract sequence representations for both the reference and mutant sequences and measure their embedding angular distance. We assess the model's ability to rank functional variants without any task-specific training by computing AUROC and AUPRC over the similarity scores.
\end{itemize}
\paragraph{Training \& Implementation Details.}
All models are implemented in PyTorch and trained within a unified JEPA-DNA framework. DNABERT-2 and NTv3 are trained on a single NVIDIA B300 GPU, using an Intel Xeon 6776P CPU with 2.0 TiB system RAM, and each model was trained for approximately five days. HyenaDNA is trained on a single NVIDIA A10 GPU, using an AMD EPYC 7R32 CPU with 200 GiB system RAM, and was trained for approximately ten days.

The context and target encoders are both initialized from pre-trained weights, while the predictor network is initialized from scratch using a truncated normal distribution with standard deviation 0.02 for all linear layer weights and zero initialization for biases.

For the masking strategy, we use a span-based scheduler that evolves over the course of training. At the beginning of training, we sample 1–2 contiguous target regions per sequence, covering 10–30\% of the sequence length. Both the number of spans and the overall masking ratio are then increased linearly (with rounding) so that, by the end of training, we sample 1–4 contiguous target regions covering 30–50\% of the sequence. The context encoder processes the masked sequence while the target encoder (updated via exponential moving average) processes the full unmasked sequence.

JEPA-DNA Continual training follows a two-phase schedule optimized for continual training and shared across all backbones. 
\begin{itemize}
\item \textbf{Phase 1 (Predictor warmup).} For the first 1{,}000 steps, the encoder is frozen and only the predictor is trained, using a learning rate of \(1 \times 10^{-5}\) for DNABERT-2 and HyenaDNA and approximately \(1.06 \times 10^{-5}\) for NTv3. This warmup allows the predictor to learn meaningful initial representations before jointly updating the encoder.
\item \textbf{Phase 2 (Full training).} The encoder is then unfrozen and trained jointly with the predictor using a 500-step linear warmup, followed by decay. For DNABERT-2 and HyenaDNA, the learning rate is linearly increased from \(3 \times 10^{-6}\) to a peak of \(5 \times 10^{-6}\) and then decayed with a cosine schedule to \(1 \times 10^{-6}\). For NTv3, we adopt a lower learning-rate regime with phase-2 learning rates of \(10^{-6}\) to \(10^{-8}\), reflecting its higher training sensitivity.
\end{itemize}

We use stochastic gradient descent (SGD)~\citep{bottou2010large} with momentum 0.9 and weight decay 0.01 for all models, with a batch size of 32 and gradient accumulation over 4 iterations (effective batch size per step is 128). Each backbone is trained for up to 100{,}000 optimization steps (each step comprising 4 gradient accumulation iterations) with an early stopping mechanism. At every checkpoint after a warmup we compute $d_{\mathrm{ema}}$, the mean $L_2$ distance between the context-encoder and EMA-target sequence embeddings on a fixed set of clean, unmasked validation sequences. Training is stopped only when (i) the previous checkpoint is a local minimum, that is, its $d_{\mathrm{ema}}$ is at most $1.15\times$ the minimum over the previous four checkpoints, and (ii) the current $d_{\mathrm{ema}}$ is at least $3\times$ that minimum, in which case we restore the local-minimum checkpoint. While training is stable the online encoder and its EMA target remain closely aligned, so $d_{\mathrm{ema}}$ stays low and a sharp rise after such a quiet period marks a genuine context--target divergence. Under this rule DNABERT-2 never triggers early stopping and runs to the full budget, whereas NTv3 and HyenaDNA stop early. The target encoder is updated via an exponential moving average (EMA) with momentum scheduled from 0.996 to 1.0 over the course of training.

The training objective combines multiple losses:
\begin{itemize}
\item Latent Predictive Loss (JEPA Loss): Cosine similarity loss between the predicted and target sequence representations, used with weight 1.0, encouraging the context encoder to capture global sequence semantics from partial observations.
\item LLM Loss: For bidirectional backbones, we apply a standard MLM objective on the masked token positions, reusing the JEPA encoder forward and masking. For autoregressive backbones, we instead use a next-token prediction (NTP) loss with an additional encoder forward under causal masking. The LLM loss term (MLM or NTP) is included with a weight of 0.1 in the total loss.
\item Variance Loss: Hinge-based variance loss with weight 25.0 to prevent representation collapse by ensuring variance above a threshold (1.0) across the batch dimension. The variance is computed on the sequence representation embeddings from both the context encoder output and the predictor output. To avoid artificial variance that does not reflect true representational diversity, we employ two strategies: (1) variance calculations are performed via an additional forward pass through the context encoder and predictor in evaluation mode, eliminating stochastic effects from dropout and random masking; and (2) all sequences within each batch are truncated to a uniform length, preventing spurious variance arising from heterogeneous padding patterns.
\item Covariance Loss: With weight 0.5 to decorrelate embedding dimensions and 
encourage diverse feature learning.
\end{itemize}

For supervised downstream evaluation, we utilize a linear probing approach where the encoder weights are frozen and a linear classification head is trained on the sequence representations. For standard single-sequence tasks, the projection layer operates directly on the sequence embedding. In variant effect prediction, the reference and variant sequences are processed independently, and their corresponding sequence embeddings are concatenated prior to classification. To ensure rigorous comparison, all supervised tasks follow a unified training protocol: the classifier is trained for 3 epochs using the AdamW optimizer with a learning rate of $3 \times 10^{-5}$, a weight decay of 0.01, and a batch size of 32. For both supervised and zero-shot tasks, when sequence length exceeds the context window of the underlying backbone, inputs are truncated to retain the central region, using 2{,}500 bp (approximately 512 BPE tokens) for DNABERT-2 and 8{,}192 bp for NTv3 and HyenaDNA.

%%%%%%%%%%%%%%%%%%%%%%%%%%%%%%%%%%%%%%%%%%%%%%%%%%%%%%%%%%%%

%\include{chapters/neurips_checklist} no need to include for arxiv
\end{document}